%% file: main.tex
\documentclass[10pt,twocolumn,letterpaper,pagenumbers]{article}

\usepackage{cvpr}              

\input{preamble}

\definecolor{cvprblue}{rgb}{0.21,0.49,0.74}
\usepackage[pagebackref,breaklinks,colorlinks,citecolor=cvprblue]{hyperref}
\usepackage{mathtools}

\def\paperID{460} 
\def\confName{3DV\xspace}
\def\confYear{2026\xspace}

\title{VI3: Grounding Pretrained 3D Foundation Models with Inertial Cues}

\author{
Ernesto Lozano \quad
Alberto Jaenal \quad
Javier Civera \\[2mm]
I3A, Universidad de Zaragoza, Spain\\
{\tt\small \{e.lozano, ajaenal, jcivera\}@unizar.es}
}

\begin{document}
\maketitle

\begin{strip}
    \centering
    \includegraphics[width=1.0\textwidth]{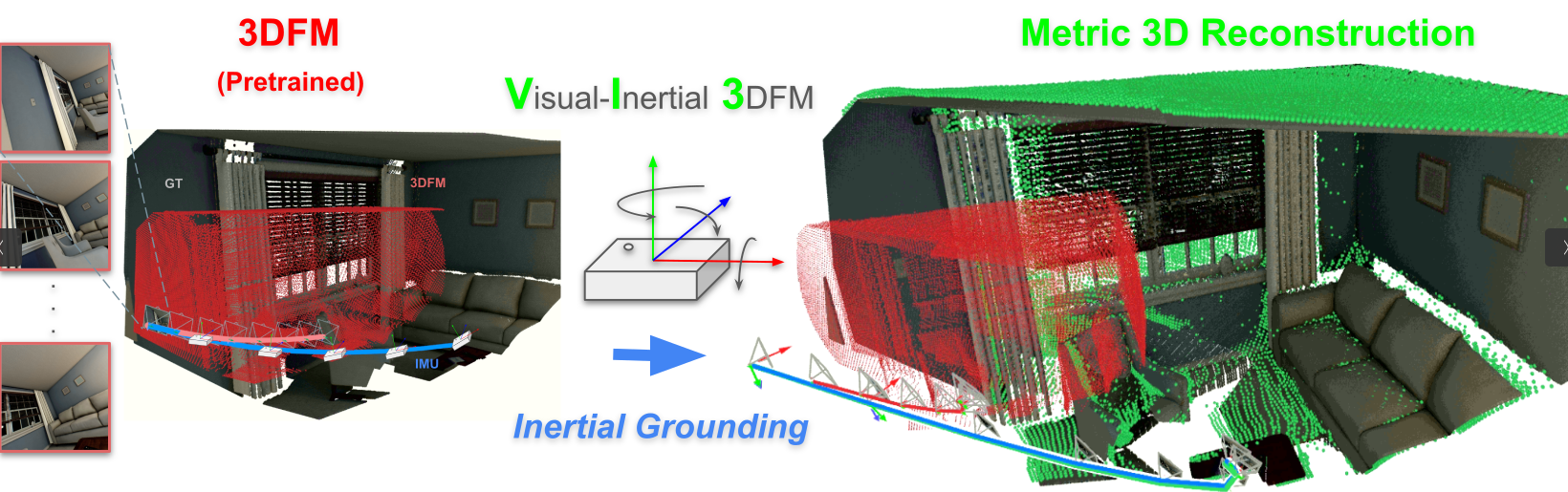}
    \captionof{figure}{VI3 grounds the predictions of a pretrained 3D foundation model (trajectory and scene reconstruction shown in red) using inertial data from an onboard IMU (inertial trajectory shown in blue) as GT-free supervision. This leads to a physics-aware, scene-specific trajectory refinement and metric scale recovery, that keeps geometry consistent as seen in the right figure (VI3 results shown in green).}
    \label{fig:teaser}
\end{strip}

\input{sec/0_abstract}    
\input{sec/1_intro}
\input{sec/2_related_work}
\input{sec/3_method}
\input{sec/4_experiments}

\input{sec/5_conclusion}

{
    \small
    \bibliographystyle{ieeenat_fullname}
    \bibliography{example,zotero}
}
\input{sec/X_suppl}

\end{document}

%% file: preamble.tex
\usepackage[dvipsnames]{xcolor}

\graphicspath{ {fig/} }
\usepackage{float}
\usepackage{stfloats}
\usepackage{placeins}
\usepackage{cuted}
\usepackage{caption}
\usepackage{xcolor}
\usepackage{algorithm,algpseudocode}
\usepackage{xfrac}
\usepackage{multirow}

%% file: sec/0_abstract.tex
\begin{abstract}
3D foundation models (3DFMs) excel at predicting camera poses and dense depth from multiple views of a scene, showcasing strong zero-shot generalization. However, as metric scale is not observable from monocular images, their absolute scale predictions are typically inaccurate.
Inertial measurement units (IMUs), present in most devices, naturally complement monocular cameras by observing scaled motion. We introduce VI3, a model-agnostic framework that metrically anchors a pretrained 3DFM using only IMU readings. VI3 initializes and preintegrates the IMU to obtain a metric motion reference, which is then used to recover the scale of the 3DFM outputs. Our method includes adaptable anchoring strategies tailored to diverse 3DFM architectures. Experiments on synthetic and real aerial datasets demonstrate that VI3 recovers metric scale without ground-truth supervision while preserving geometric consistency, acting as a fine refinement under well-conditioned motion and as a strong prior when motion is less informative.
\url{https://ernestolozano.github.io/vi3}
\end{abstract}

%% file: sec/1_intro.tex
\section{Introduction}
\label{sec:intro}


Estimating 3D scene attributes from unstructured images is a cornerstone of computer vision, with applications in robotics~\cite{mahdavian2026uniscale} and neural rendering~\cite{wang2026nerfs} among others.
For decades, the prevailing paradigm has relied on feature-based pipelines, \eg Structure-from-Motion (SfM)~\cite{agarwal2011building,schoenberger2016sfm, pan2024global} or visual SLAM~\cite{Mur_Artal_2015_orbslam,mur2017orb,keetha2024splatam}. Despite their trajectory accuracy, these methods remain brittle in unconstrained environments~\cite{sattler2017large, panek2026guide}, yielding sparse point clouds rather than the dense geometry required for modern spatial AI~\cite{bao20253d,wang2026nerfs}. Recently, 3D foundation models (3DFMs) have reshaped this landscape. Trained on large, diverse image collections, models such as DUSt3R~\cite{wang_dust3r_2023}, VGGT~\cite{wang_vggt_2025}, $\pi$³~\cite{wang_pi3_2025}, and a growing list of successors~\cite{leroy_master_2024, lin_depthA3_2025, keetha_mapanything_2026, wang_vggt-_2026, peng2026omnivggt} learn strong geometry and motion priors, predicting camera poses and dense per-pixel geometry from unposed images in a single feed-forward pass.


Nevertheless, 3DFMs process images independently through monocular backbones~\cite{oquab_dinov2_2024}. While this enables them to preserve highly accurate relative motion and scene structure, their reconstructions remain defined only up to an unknown scale factor and are therefore decoupled from real-world metric dimensions. This limitation derives from their training: 3DFMs learn under scale-invariant supervision objectives (usually distilled from SfM pipelines~\cite{schoenberger2016sfm} rather than metric signals), making them fundamentally incapable of determining the metric scale of the scene.
This remains an open gap: turning a scale-free prediction into a physically anchored, metric reconstruction.
Recent work have attempted to enforce metric scale by conditioning inference using camera priors~\cite{jang_pow3r_2025,mahdavian2026uniscale} or through dedicated tokens~\cite{keetha_mapanything_2026}. However, their strong data-driven priors may override explicit metric signals, yielding reconstructions that fail to reflect the real-world dimensions. Crucially, such methods still depend on precise camera calibration (often unavailable in unconstrained settings) or upstream metric poses to produce physically-anchored reconstructions.

This gap can be fully addressed by aligning vision with physical measurements from a sensor ubiquitous across most platforms: an inertial measurement unit (IMU). By providing metric readings, the IMU renders real-world scale observable, with a decade of visual-inertial research~\cite{forster_-manifold_2017,Geneva2020ICRAopenvins,campos_orb-slam3_2021} sustaining its sufficiency to metrically ground vision.


We introduce a model-agnostic \textbf{V}isual \textbf{I}nertial grounding for \textbf{3}DFMs (VI3) that enables metric 3DFM reconstruction by just using an IMU; entirely bypassing optimization backends, maps, or external finetuning corpora. Given a short clip, the IMU is initialized leveraging the outputs of the 3DFM (without requiring any GT or external signal) and then preintegrated into a physics-consistent, gravity-aware motion reference that anchors the 3DFM to the real-world. Due to the diversity of 3DFM models, we propose different ways of injecting the metric scale, such as input conditioning (for compatible modes) or test-time refinement~\cite{han2025vittt}, thereby scaling the predicted geometry without explicit 3D supervision.


In summary, our contributions are threefold. (1)~To the best of our knowledge, VI3 is the first method to bring together at inference time the dense 3DFM priors and the metric grounding of inertial sensing with the aim of physically anchoring any model, only requiring an on-board IMU. (2)~We propose an IMU intialization method that uses the 3DFM and the inertial readings to estimate the IMU state. (3)~We present a test-time refinement formulation to inject physical awareness into 3DFMs and render metric-scale reconstructions. Through extensive experiments on diverse aerial datasets and across several representative 3DFM baselines, we show that VI3 metrically grounds a frozen 3DFM from a coarse inertial motion estimate, yielding reconstruction scales close to the real-world. 

%% file: sec/2_related_work.tex
\section{Related Work}
\label{sec:relwork}

\noindent\textbf{Feedforward 3D reconstruction.}
3DFMs replace the iterative correspondence-and-optimization loop of classical pipelines~\cite{mur2017orb,schoenberger2016sfm}, by regressing camera poses and dense per-pixel geometry from unposed images in a feedforward pass. DUSt3R~\cite{wang_dust3r_2023} and MASt3R~\cite{leroy_master_2024} predict pointmaps from image pairs, with a family of successors focusing on incremental reconstructions~\cite{wang2025spann3r,cabon2025must3r,cut3r,liu2025slam3r,zhuo2025streamvggt,chen2025long3r,lan2025stream3r,li2025wint3r,wu2025point3r,ma2025puzzles,mahdi2025evict3r,chen_ttt3r_2026}; while VGGT~\cite{wang_vggt_2025}, $\pi^3$~\cite{wang_pi3_2025}, and subsequent models~\cite{lin_depthA3_2025, wang_vggt-_2026, tang2025mvdust3r,elflein2025light3rsfm,yang2025fast3r} extend this to multiple views and estimate a set of common 3D attributes jointly (e.g. depth, cameras, pointmaps). These models provide strong and dense geometric priors but are trained under SfM labels, predicting geometry only up to an unknown global scale. 

\noindent\textbf{Scale-aware 3DFMs.}
Recent works have attempted to inject metric scale into 3DFM reconstructions. Some works~\cite{miyato2024gta,li2026cameras} integrate it directly into the attention mechanism. Other approaches opt to use modular auxiliary heads to ingest priors (intrinsics, poses, metric depth) to condition the network during inference. AMB3R~\cite{wang2026amb3r} uses a head specifically trained to regress a data-driven scale from the model features. In turn, Pi3X~\cite{wang_pi3_2025}, Depth Anything 3~\cite{lin_depthA3_2025}  and others~\cite{jang_pow3r_2025,khafizov2025g,mahdavian2026uniscale} directly produce metric reconstructions from the scaled inputs, while MapAnything~\cite{keetha_mapanything_2026} and variants~\cite{karhade2026any4d} use a scale token that allows them to regress the metric scale factor. However, the prior metric scale is still learned, inherited from the training distribution rather than measured; while any metric input depends on a dedicated frontend application. We propose a model-agnostic framework that physically anchors any 3DFM only using inertial signals; conditioning or refining the model output.

\noindent\textbf{Visual-inertial estimation.}
Inertial sensing resolves the scale ambiguity of monocular vision. Classical visual-inertial systems~\cite{qin2018vins,Geneva2020ICRAopenvins,campos_orb-slam3_2021,boche2025okvis2} estimate metric, gravity-aligned trajectories in real time, but their reconstruction is sparse, covering only trackable features, and runs on a stateful backend tuned to a fixed rig. Closer to our approach, recent work couples unconstrained, learned 3D with inertial data: MASt3R-Fusion~\cite{zhou2025mast3rfusion} fuses feedforward pointmap correspondences with IMU preintegration in a factor-graph backend, and VIGS-SLAM~\cite{ZHU2026VIGS-SLAM} optimizes an explicit per-scene 3D Gaussian representation jointly with the IMU states. In both, the inertial signal drives an external optimization, and the pretrained weights are never adapted. This work builds no explicit representation nor recurrent state, instead we use the inertial signal as a lightweight geometric anchor to condition or adapt the model output. 

\noindent\textbf{Test-time adaptation of 3D feedforward models.}
To avoid costly retraining, test-time adaptation updates the weights of the pretrained foundation model during inference to meet a target objective. Generic vision test-time training~\cite{han2025vittt} tunes parameters to each test input to improve robustness under distribution shifts. For 3DFM reconstruction, recent work~\cite{cut3r, chen_ttt3r_2026, zhang2026loger, jin2026zipmap} focuses on refinement for long sequence stability and affordability, to achieve linear-time, drift-free visual and temporal consistency. We propose refinement as a way to adapt the 3DFM weights by injecting metric scale to a short monocular video, anchoring the resulting geometry to a physical reference.

%% file: sec/3_method.tex
\section{Up-to-Scale Geometry from 3DFMs}
\label{sec:method}

\begin{figure*}[!t]
    \centering
    \includegraphics[width=1.0\textwidth]{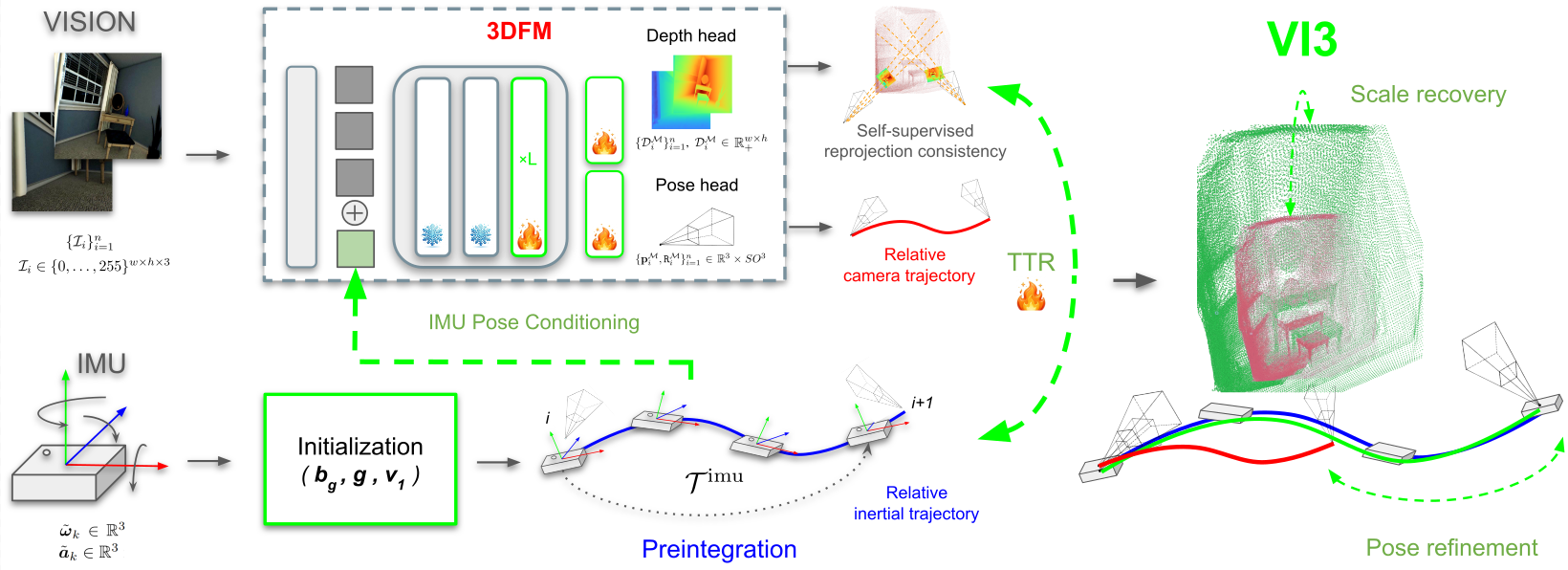}
    \caption{VI3's block diagram, consisting of two branches. \emph{Visual:} a pretrained 3DFM takes unposed input images and yields depth and poses. \emph{Inertial:} the on-board IMU stream, gt-free initialized and preintegrated, provides a scene-specific target trajectory. On-board inertial cues (green dashed arrows), introduced either via pose conditioning, or through adaption (TTR), ground the reconstruction metric.}
    \label{fig:block_diagram}
\end{figure*}

A 3DFM $\mathcal{M}$ receives as input $n$ unposed consecutive frames $\{\mathcal{I}_i\}_{i=1}^n$, with ${\mathcal{I}_i \in \{0,\hdots,255\}^{w \times h \times 3}}$, and predicts a set of geometric outputs 
that depends on the specific model, but typically includes pixel-wise depth $\{\mathcal{D}^\mathcal{M}_i \}_{i=1}^n$, $\mathcal{D}^\mathcal{M}_i \in \mathbb{R}_{+}^{w \times h}$ and camera extrinsics $ \{\mathbf{p}_i^\mathcal{M}, \mathtt{R}_i^\mathcal{M} \}_{i=1}^n \in \mathbb{R}^3 \times SO^3$ for each input image. We assume that camera intrinsics are pre-calibrated, although 3DFMs can also predict them. Point cloud representations ${\mathcal{X}^\mathcal{M}_i \in \mathbb{R}^{w \times h \times 3}}$ can be computed per image from the depth maps and intrinsics.

\section{Inertial Preintegration}

An IMU provides at a discrete time instant $k$ measurements of the angular velocity $\tilde{\boldsymbol{\omega}}_k \in \mathbb{R}^3$ and linear acceleration $\tilde{\boldsymbol{a}}_k \in \mathbb{R}^3$ in the local reference frame, which are affected by additive, slowly-varying biases $\boldsymbol{b}^g_k \in \mathbb{R}^3$ and $\boldsymbol{b}^a_k \in \mathbb{R}^3$ and and zero-mean Gaussian noises ${\boldsymbol{\eta}^{g} \sim \mathcal{N}(\boldsymbol{0},\boldsymbol{\Sigma}_g)} \in \mathbb{R}^3$ and ${\boldsymbol{\eta}^{a} \sim \mathcal{N}(\boldsymbol{0},\boldsymbol{\Sigma}_a)} \in \mathbb{R}^3$
\begin{equation}
    \boldsymbol{\omega}_k = \tilde{\boldsymbol{\omega}}_k - \boldsymbol{b}^g_k-\boldsymbol{\eta}^{g},  \qquad
    \boldsymbol{a}_k = \tilde{\boldsymbol{a}}_k-\boldsymbol{b}^a_k-\boldsymbol{\eta}^{a},
\end{equation}
where $\boldsymbol{\omega}_k$ and $\boldsymbol{a}_k$ refer to the true values for angular velocity and acceleration, respectively. As we work with short visual-inertial \emph{clips}, we will assume constant biases  in the considered intervals, \emph{i.e.}, $\boldsymbol{b}^g_k \approx \boldsymbol{b}^g$ and $\boldsymbol{b}^a_k \approx \boldsymbol{b}^a$.

\subsection{Preintegration and state propagation}
\label{sec:preint}

Preintegrating~\cite{forster_-manifold_2017} the $m$ bias- and noise-corrected IMU readings between two consecutive camera frames, $i$ and $i+1$, yields the relative rotation $\Delta \mathtt{R}_i = \mathtt{R}_i^{\top}\mathtt{R}_{i+1}$, relative position $\Delta \mathbf{p}_i = \mathbf{p}_{i+1} - \mathbf{p}_i$ and velocity change $\Delta \mathbf{v}_i = \mathbf{v}_{i+1} - \mathbf{v}_i$ between them:

\begin{align}
	\Delta \mathtt{R}_i
    =&
    \prod_{k=1}^{m} \mathrm{Exp} \left(\left(\tilde{\boldsymbol{\omega}}_k  \!-\! \boldsymbol{b}^g \!-\!\boldsymbol{\eta}^{g} \right) \Delta t\right),  \label{eq:preintegrationRotation}\\
  \Delta \mathbf{v}_i
    =& \ \mathbf{g} \Delta T
    + \sum_{k=1}^{m} \mathtt{R}_k \Big( \tilde{\boldsymbol{a}}_k-\boldsymbol{b}^a-\boldsymbol{\eta}^{a} \Big)\Delta t,  \label{eq:preintegrationVelocity}\\ 
  \Delta \mathbf{p}_i
    =&\ \sum_{k=1}^{m}
      \Big[ \mathbf{v}_k \Delta t
    + \frac{1}{2}\mathbf{g} \Delta t^2
    + \frac{1}{2}\mathtt{R}_k \Big( \tilde{\boldsymbol{a}}_k \!-\! \boldsymbol{b}^a \!-\!\boldsymbol{\eta}^{a} \Big)\Delta t^2 \Big] \label{eq:preintegrationTranslation}
\end{align}

\noindent where $\Delta t$ stands for the time between IMU readings, ${\Delta T = m \Delta t}$ for the time between consecutive frames, $\mathrm{Exp}$ for the exponential map from $\mathfrak{so}(3)$ to $SO(3)$, and $\mathbf{g}$ for the gravity vector. 

Our goal is to recover the inertial states $\mathtt{R}_i$, $\mathbf{p}_i$ and $\mathbf{v}_i$ for every frame. We will assume the origin at the first frame, hence $\mathtt{R}_1 = \mathtt{I}_3$ and $\mathbf{p}_1 = \mathbf{0}$, and the states will be recoverable if we know the values for the biases $\boldsymbol{b}^g$ and $\boldsymbol{b}^a$, the gravity $\mathbf{g}$ and the velocity in the first frame $\mathbf{v}_1$.

\subsection{Inertial state initialization}

\label{sec:init}

As shown in by \citet{kaiser2016simultaneous}, the effect of $\boldsymbol{b}^a$ is negligible for short-time integrations, so we will assume $\boldsymbol{b}^a \approx \mathbf{0}$ and will only estimate the values for $\boldsymbol{b}^g$, $\mathbf{v}_1$ and $\mathbf{g}$ from the up-to-scale geometry estimated by the 3DFM and the IMU readings.

\paragraph{Gyroscope bias ($\boldsymbol{b}^g$).}
As observed in Equation~\ref{eq:preintegrationRotation}, $\mathbf{b}_g$ can be estimated from the angular velocity measurements given an estimate of the relative rotation $\Delta \mathtt{R}_i$, which we obtain from the 3DFM. We adopt the closed-form solution by \citet{cerezo2026efficient}, which, assuming small-angle rotations, estimates $\mathbf{b}_g$ from two consecutive frames $\mathcal{I}_i$ and $\mathcal{I}_{i+1}$ as
\begin{equation}
{\boldsymbol{b}}^g_i\approx-\tfrac{1}{\Delta t}
\operatorname{Log}\!\Big(\operatorname{Exp}(\bar{\boldsymbol{\omega}}_{i}\Delta t)^{\!\top}
\operatorname{Exp}\big(\tfrac{1}{m}\operatorname{Log}\Delta\mathtt{R}^{\mathcal{M}}_{i}\big)\Big),
\label{eq:bg}
\end{equation}
where $\bar{\boldsymbol{\omega}}_{i}=\tfrac{1}{m}\sum_{k=1}^{m}\tilde{\boldsymbol{\omega}}_k$ is the average of angular velocity measurements between the two frames, $\operatorname{Log}$ is the logarithmic map from $SO(3)$ to $\mathfrak{so}(3)$, and $\Delta\mathbf{R}^{\mathcal{M}}_{i} = \mathbf{R}{^{\mathcal{M}}_{i}}^\top \mathbf{R}{^{\mathcal{M}}_{i+1}}$ is the relative rotation estimated by the 3DFM. The final estimate for $\hat{\boldsymbol{b}}^g$ is obtained averaging the values for all $n-1$ pairs over the sequence $\hat{\boldsymbol{b}}^g = \tfrac{1}{n-1}\sum_{i=1}^{n-1}{\boldsymbol{b}}^g_i$

\paragraph{Initial velocity ($\mathbf{v}_1$).} Up-to-scale velocities can be computed from the 3DFM poses using discrete differences as ${\mathbf{v}^{\mathcal{M}}_{i} = \sfrac{\Delta \mathbf{p}_i^{\mathcal{M}}}{\Delta T}}$. Assuming for now known gravity $\mathbf{g}$, something that we will resolve in the next paragraphs, the 3DFM scale factor $s$ can be constrained using the velocity increments in Eq.~\ref{eq:preintegrationVelocity}, yielding ${s \Delta \mathbf{v}^{\mathcal{M}}_{i} = \Delta \mathbf{v}_i}$, where $\Delta \mathbf{v}^{\mathcal{M}}_{i} = \mathbf{v}^{\mathcal{M}}_{i+1} - \mathbf{v}^{\mathcal{M}}_{i}$. Stacking these constraints over all $n-1$ image pairs gives the linear system ${s \mathbf{V}^{\mathcal{M}} =  \mathbf{V}}$, with $\mathbf{V}^{\mathcal{M}} = \begin{pmatrix} {\Delta \mathbf{v}^{\mathcal{M}}_{1}}^\top & \hdots & {\Delta \mathbf{v}^{\mathcal{M}}_{n-1}}^\top \end{pmatrix}^\top$ and ${\mathbf{V} = \begin{pmatrix} {\Delta \mathbf{v}_{1}}^\top & \hdots & {\Delta \mathbf{v}_{n-1}}^\top \end{pmatrix}^\top}$. Since both velocity estimates are noisy, we formulate the problem as a total least squares (TLS) optimization rather than ordinary least squares, and obtain an initial estimate $\hat{s}$ by solving
\begin{align}
\hat{s}_{\text{init}} &= \arg\min_{s,\Delta \mathbf{V}^{\mathcal{M}},\Delta \mathbf{V}} || \Delta \mathbf{V}^{\mathcal{M}} \ \ \Delta \mathbf{V} ||_F^2 \\
& \text{s.t.} \ \ \ s (\mathbf{V}^{\mathcal{M}} + \Delta \mathbf{V}^{\mathcal{M}}) = \mathbf{V} + \Delta \mathbf{V}\nonumber
\end{align}
which, in this one-dimensional case, has the following closed-form solution
\begin{equation}
\hat{s}=\operatorname{sign}\big(\langle\mathbf{V},\mathbf{V}^{\mathcal{M}}\rangle\big)
\frac{\|\mathbf{V}\|}
{\|\mathbf{V}^{\mathcal{M}}\|}.
\label{eq:S}
\end{equation}
As this initial estimation of the scale may be affected by large errors in the velocity estimations, in particular for low-parallax frames, we refine this estimation as follows.
We align the initial velocity's orientation ${\breve{\mathbf{v}}_1={\mathbf{v}}^{\mathcal{M}}_1/\|{\mathbf{v}}^{\mathcal{M}}_1\|}$ with the gravity-free first pair, and we compute its magnitude $\mu$ as the median of back-projecting all frames:
\begin{equation}
\mu=\operatorname{median}_k\big\langle \hat{s}{\mathbf{v}}^\mathcal{M}_k
-\textstyle\sum_{1<l\leq k}\Delta\mathbf{v}_l,\;\hat{\mathbf{v}}_1\big\rangle
\label{eq:v0}
\end{equation}
The initial velocity is finally computed as ${\mathbf{v}_1=\mu\,\breve{\mathbf{v}}_1}$. Two sanity checks safeguard the optimization: $s$ falls back to $1$ if visual and inertial motions are anti-aligned, and $\mathbf{v}_1$ reverts to $s{\mathbf{v}}^\mathcal{M}_1$ if $\mu\le 0$ due to an inverted gravity vector.

\paragraph{Gravity ($\mathbf{g}$).}
The gravity vector can be estimated from image cues using GeoCalib~\cite{veicht2024geocalib}, yielding an estimate that we will denote as $\mathbf{g}_{\mathrm{gc}}$. However, we observed that these estimates can degrade severely on poorly textured, out-of-distribution images, while GeoCalib's confidence measure does not reliably reflect these large errors. To detect such failure cases, we compute approximate gravity estimates using two additional methods and assess their mutual consensus. 

Assuming negligible accelerometer bias and small linear accelerations, we can obtain a coarse gravity estimate $\mathbf{g}_{\text{imu}}$ by averaging the measured accelerations
\begin{equation}
  \mathbf{g}_{\mathrm{imu}}
  = -9.81\frac{\sum\nolimits_{k=1}^{(n-1)m}
  \mathtt{R}_k\tilde{\mathbf{a}}_k}{(n-1)m\lVert \sum\nolimits_{k=1}^{(n-1)m}
  \mathtt{R}_k\tilde{\mathbf{a}}_k \rVert}
  \label{eq:gimu}
\end{equation}

Additionally, we can also derive the gravity with a kinematic least-squares procedure, an alternative that we will denote as $\mathbf{g}_{\mathrm{ls}}$. Based on the preintegrated velocity update~\cite{forster_-manifold_2017} across frame pair $(i, j)$:
\begin{equation}
  \mathbf{v}_j = \mathbf{v}_i + \mathbf{g}\,\Delta t_{ij}
  + \!\!\sum_{i<k\leq j}\!\! \mathbf{R}_{k-1}\Delta\mathbf{v}_k,
  \label{eq:preint}
\end{equation}
we substitute the difference between the visual observation scaled to metric units $\mathbf{v}_j-\mathbf{v}_i=s\,\Delta{\mathbf{v}}^\mathcal{M}_{ij}$, isolating the velocity residuals in the integrated inertial motion:

\begin{equation}
  \mathbf{r}_{ij}
  \;\coloneqq\; s\,\Delta{\mathbf{v}}^\mathcal{M}_{ij}
  - \!\!\sum_{i<k\leq j}\!\! \mathbf{R}_{k-1}\Delta\mathbf{v}_k
  \;=\; \mathbf{g}\,\Delta t_{ij},
  \label{eq:resid}
\end{equation}

Minimizing the squared norm $\sum_{i<j}\Vert{}\Delta t_{ij}\mathbf{g} - \mathbf{r}_{ij}\Vert{}^2$ over all image pairs $i<j$ yields a closed-form solution for the unit gravity direction:

\begin{equation}
  \mathbf{g}_{\mathrm{ls}}
  = 9.81,\frac{\mathbf{n}}{\lVert\mathbf{n}\rVert},
  \ \ 
  \mathbf{n} = \frac{\sum_{i<j}\Delta t_{ij}\,\mathbf{r}_{ij}}
                    {\sum_{i<j}\Delta t_{ij}^{2}}
  = \sum_{i<j} w_{ij}\,\frac{\mathbf{r}_{ij}}{\Delta t_{ij}},
  \label{eq:gls}
\end{equation}

where the weighting term $w_{ij}=\Delta t_{ij}^{2}/\sum\Delta t_{ij}^{2}$ makes long-baseline pairs dominate quadratically, insulating $\mathbf{g}_{\mathrm{ls}}$ against high-frequency visual tracking noise while locking the magnitude to the gravity modulus $9.81$.
There is a circular dependence between the term $\mathbf{g}_{\mathrm{ls}}$ and the scale $s$, that was mentioned when estimating the initial velocity $\mathbf{v}_1$, which can be modeled as the function ${s=\mathcal{S}(\mathbf{g})}$. This requires to tightly couple both variables through a fixed-point iteration scheme $  {s^{(n+1)}=\mathcal{S}\big(\mathbf{g}^{(n)}\big),}\,\,{\mathbf{g}^{(n+1)}=\mathbf{g}_{\mathrm{ls}}\big(s^{(n+1)}\big)}$, which starts from a scale-free seed $\mathbf{g}^{(0)}$ ($\mathbf{g}_{\mathrm{gc}}$ or $\mathbf{g}_{\mathrm{imu}}$) and converges within two to three iterations to solution $\mathrm{FP}(\mathbf{g}^{(0)})$.

The consensus is evaluated using the two independent seeds for the iterative process: visual ${\hat{\mathbf{g}}_{\mathrm{gc}} = \mathrm{FP}(\mathbf{g}_{\mathrm{gc}})}$ and inertial ${\hat{\mathbf{g}}_{\mathrm{imu}} = \mathrm{FP}(\mathbf{g}_{\mathrm{imu}})}$. 
When the motion is sufficient, both seeds converge to a common direction within approximately $1^\circ$, establishing a consensus error $\angle(\hat{\mathbf{g}}_{\mathrm{gc}}, \hat{\mathbf{g}}_{\mathrm{imu}})$  that reflects how inertial observations pin gravity, and demonstrating that $\mathbf{g}_{\mathrm{gc}}$ is accurate in most cases.

We classify $\mathbf{g}_{\mathrm{gc}}$ as an outlier if it diverges beyond the consensus error, enforcing the condition:
\begin{equation}
  \min_{X\in\{\mathrm{gc},\mathrm{imu}\}} \angle\big(\mathbf{g}_{\mathrm{gc}},\mathbf{g}_X\big)
  \;>\;
  \max\big(\angle(\hat{\mathbf{g}}_\mathrm{gc},\hat{\mathbf{g}}_\mathrm{imu}),\,\gamma_{\min}\big),
  \label{eq:guard}
\end{equation}
where $\gamma_{\min} = 20^\circ$ acts as a lower noise floor. When this condition triggers, the system rejects $\mathbf{g}_{\mathrm{gc}}$ and adopts the purely inertial fixed point $\hat{\mathbf{g}}_\mathrm{imu}$, subject to (${\operatorname{sign}(\hat{\mathbf{g}}_\mathrm{imu})\geq0}$), alongside its corresponding scale $s_\mathrm{imu}$ and initial velocity $\mathbf{v}_1(\mathbf{g}_\mathrm{imu}, s_\mathrm{imu})$, ensuring robust initialization even under severe visual prior degradation.

Once the initialization parameters are estimated, we use the in the preintegration process described in \cref{sec:preint} to construct the metric, frame-1-anchored target trajectory $\mathcal{T}^\mathrm{imu} = {\{\mathbf{p}_i^{\mathrm{imu}}, \mathtt{R}_i^{\mathrm{imu}}\}}_{i=1}^n$ used as supervisory signal driving the anchoring.

\input{tables/non_metric}

\section{Metrically anchoring 3DFMs}

VI3 aims to inject the metric information encoded in the inertial trajectory into the pretrained 3DFM at test time. We consider two possible approaches to achieve this.

\noindent\textbf{Input conditioning} As stated in \cref{sec:relwork}, certain models~\cite{wang_pi3_2025,lin_depthA3_2025} can use optional inputs during inference to metrically scale the output. Thus, we feed $\mathcal{T}^\mathrm{imu}$ into the 3DFM to obtain a physically-informed reconstruction.

\noindent\textbf{Test-time refinement} The objective of the proposed test-time refinement process is to find a joint parameter subset $(\alpha, \hat{\theta})$ of scale and model weights such that the geometric output of the refined model is physically aligned to the trajectory $\mathcal{T}^\mathrm{imu}$.
Since the geometry predicted by the 3DFM is correct up to a global scale factor, we introduce a single learnable scalar $\alpha > 0$, which from onwards acts on the raw translation ${\mathbf{p}^\mathcal{M}_i \mapsto \alpha\,\mathbf{p}^\mathcal{M}_i}$ and depth ${\mathcal{D}^\mathcal{M}_i \mapsto \alpha\,\mathcal{D}^\mathcal{M}_i}$, tying both representations and leaving rotation and intrinsics untouched. This factor is initialized by directly comparing the ratios between the path lengths $
\alpha_0 = \Big( \sum_i \| \mathbf{p}_i^{\mathrm{imu}} \|_2 \Big) \Big/ \Big( \sum_i \| {\mathbf{p}^\mathcal{M}}_i \|_2 \Big)$.

We propose a pose supervision term that takes place on consecutive frames, making the objective globally invariant (rigid transforms shared by prediction and target cancel):
\begin{align}
\mathcal{L}_{\mathrm{pos}} &= \frac{1}{n-1} \sum_{i=2}^{n}
  \big\| ({\mathbf{p}}^\mathcal{M}_i - {\mathbf{p}}^\mathcal{M}_{i-1})
       - \Delta \mathbf{p}_{i-1} \big\|_1, \\
                    \mathcal{L}_{\mathrm{rot}}   &= \frac{1}{n-1} \sum_{i=2}^{n}
 \big\|\operatorname{Log}\left( \left( {{\mathtt{R}}^\mathcal{M}_{i-1}}^\top {\mathtt{R}}^\mathcal{M}_i\right)^\top
                    \Delta \mathtt{R}_{i-1}\right)\big\|_2.
\end{align}

Since this procedure operates without an available 3D ground truth, we need to enforce cross-view geometric consistency by coupling depth maps and camera poses via a self-supervised consistency loop. 
We thus introduce a geometric depth-consistency loss that anchors predicted depths directly to the inertial recovered pose scale. For every pair $(i,j)$ where $i \neq j$, we penalize discrepancies between the reprojected depth and the source depth sampled at the valid reprojected coordinates~\cite{monodepth2}: 
\begin{equation}
\small
\mathcal{L}_{\mathrm{dcons}} = \frac{1}{n} \sum_{t=1}^{n} \Big[\, \min_{i \neq j} \Big| [\pi_i(\mathcal{X}^\mathcal{M}_j)]_Z - \mathcal{D}^\mathcal{M}_i\!\left( \pi_i(\mathcal{X}^\mathcal{M}_j) \right) \Big| \,\Big]_{\mathrm{valid}},
\label{eq:dcons}
\end{equation}
with $\pi_i$ the projection operator projecting into the $i$-th frame the 3D world points $\mathcal{X}^\mathcal{M}_j$ resulting of backprojecting depth predictions from frame $j$; and $[\,]_Z$ the z-coordinate.

On top of these, we add two regularization terms that protect the structural integrity: (i) a frame-$1$ positional anchor on the first frame predicted position $\hat{\mathbf{p}}^\mathcal{M}_1$, restricting the coordinate origin ($\mathcal{L}_{\mathrm{a}}^p = \| \hat{\mathbf{p}}^\mathcal{M}_1 \|_1$), and (ii) a depth-to-init anchor, which pins the depth to its zero-shot shape $
\mathcal{L}_{\mathrm{a}}^{\mathcal{D}} = \frac{1}{n} \sum_i \| \mathcal{D}^\mathcal{M}_i - {\mathcal{D}^\mathcal{M}_i}^{(0)} \|_1
$. 
This allows the network to adapt to the input stream photometric distribution while ensuring the geometry remains strictly bounded by the 3DFM structural prior. The total objective amounts to:
\begin{align}
\mathcal{L}(\theta) =
  w_r \mathcal{L}_{\mathrm{rot}} + w_p \mathcal{L}_{\mathrm{pos}}
  + w_{\mathrm{d}} \mathcal{L}_{\mathrm{dcons}}
  + w_{a_p }\mathcal{L}_{\mathrm{a}}^p
  + w_{a_d }\mathcal{L}_{\mathrm{a}}^{\mathcal{D}}
\end{align}
which we minimize by unfreezing only 
the camera- and depth-estimation heads, together with the last few $L$ inter-frame (global) attention blocks of the aggregator in the 3DFM, plus the scalar $\alpha$, running $K$ gradient steps on the single window with $\mathcal{T}^\mathrm{imu}$ held fixed.

%% file: tables/non_metric.tex
\begin{table*}[t]
\centering
\normalsize
\caption{\textbf{Up-to-scale models.} GT-free inertial test-time refinement (\textbf{TTR}) of the pretrained up-to-scale 3D foundation models (\textbf{Base}), which
predict geometry without metric scale. The base models are $\sim$3--7$\times$ under-scaled and thus score $\approx$0 in $\delta{<}1.25$,
while TTR restores absolute metric depth and pulls both trajectory and depth scale
toward $1$, respecting the geometry. Note, in addition, that rotation metrics are also substantially improved. }
\label{tab:non-metric}
\begin{tabular}{@{}c l *{12}{c}@{}}
\toprule
& \multirow{2}{*}{Model} & \multicolumn{2}{c}{Trans (m) $\downarrow$} & \multicolumn{2}{c}{Rot ($^\circ$) $\downarrow$} & \multicolumn{2}{c}{{\footnotesize Scale {\scriptsize(Traj.)} $\rightarrow$1}} & \multicolumn{2}{c}{AbsRel $\downarrow$} & \multicolumn{2}{c}{$\delta{<}1.25$ $\uparrow$} & \multicolumn{2}{c}{{\footnotesize Scale ({\scriptsize Depth}) $\rightarrow$1}} \\
\cmidrule(lr){3-4}\cmidrule(lr){5-6}\cmidrule(lr){7-8}\cmidrule(lr){9-10}\cmidrule(lr){11-12}\cmidrule(lr){13-14}
& & Base & TTR & Base & TTR & Base & TTR & Base & TTR & Base & TTR & Base & TTR \\
\midrule
\multirow{3}{*}{\rotatebox[origin=c]{90}{\scriptsize\emph{TartanAirV2}}}
  & VGGT & 1.130 & \textbf{0.358} & 0.933 & \textbf{0.410} & 0.14 & \textbf{1.05} & \textbf{0.052} & 0.089 & 0.000 & \textbf{0.515} & 0.17 & \textbf{1.11} \\
  & VGGT-$\Omega$ & 1.110 & \textbf{0.378} & 0.725 & \textbf{0.354} & 0.17 & \textbf{1.10} & \textbf{0.033} & 0.080 & 0.000 & \textbf{0.596} & 0.19 & \textbf{1.12} \\
  & $\pi$³ & 1.047 & \textbf{0.350} & 0.560 & \textbf{0.296} & 0.20 & \textbf{1.09} & \textbf{0.041} & 0.083 & 0.000 & \textbf{0.556} & 0.21 & \textbf{1.12} \\
\midrule
\multirow{3}{*}{\rotatebox[origin=c]{90}{\scriptsize\emph{EuRoC}}}
  & VGGT & 0.192 & \textbf{0.061} & 0.293 & \textbf{0.285} & 0.27 & \textbf{0.98} & 0.239 & \textbf{0.234} & 0.004 & \textbf{0.324} & 0.28 & \textbf{1.00} \\
  & VGGT-$\Omega$ & 0.176 & \textbf{0.074} & 0.291 & \textbf{0.252} & 0.32 & \textbf{1.05} & 0.212 & \textbf{0.215} & 0.003 & \textbf{0.243} & 0.31 & \textbf{1.10} \\
  & $\pi$³ & 0.178 & \textbf{0.078} & 0.234 & \textbf{0.164} & 0.40 & \textbf{1.08} & 0.204 & \textbf{0.203} & 0.008 & \textbf{0.251} & 0.42 & \textbf{1.18} \\
\midrule
\multirow{3}{*}{\rotatebox[origin=c]{90}{\scriptsize\emph{UZH-FPV}}}
  & VGGT & 1.109 & \textbf{0.497} & 2.152 & \textbf{0.301} & 0.22 & \textbf{0.81} & 0.209 & \textbf{0.203} & 0.002 & \textbf{0.201} & 0.35 & \textbf{0.92} \\
  & VGGT-$\Omega$ & 1.089 & \textbf{0.328} & 0.762 & \textbf{0.192} & 0.20 & \textbf{0.92} & 0.244 & \textbf{0.237} & 0.007 & \textbf{0.217} & 0.31 & \textbf{1.20} \\
  & $\pi$³ & 0.891 & \textbf{0.396} & 0.734 & \textbf{0.246} & 0.29 & \textbf{0.79} & \textbf{0.241} & 0.249 & 0.039 & \textbf{0.138} & 0.45 & \textbf{1.07} \\
\bottomrule
\end{tabular}
\end{table*}

%% file: sec/4_experiments.tex
\section{Experiments}
\label{sec:experiments}

\input{tables/metric}

\input{tables/calib}
\label{tab:imu_init}
\input{tables/init}

We evaluate the performance of VI3 through quantitative and qualitative results, and ablating design choices.

\noindent\textbf{Datasets.} We use three visual-inertial aerial datasets spanning synthetic, micro-aerial and drone racing nature. \emph{TartanAirV2}~\cite{wang_tartanair_2020} contributes $24$ synthetic drone-flight sequences coming from four varied environments: indoors and outdoors (House, Hospital, ModernCityDowntown, ForestEnv) at the two existing difficulties (easy/hard) and on the first two trajectories (P000, P002). We take the full \emph{EuRoC-MAV}~\cite{euroc-mav} dataset, consisting of $11$ sequences: Vicon-Room (GT-depth obtained from Leica MS50 laser scanner) and Machine-Hall (GT-depth stereo-derived). \emph{UZH-FPV}~\cite{uzh-fpv} brings $16$ drone-racing sequences with public GT (Snapdragon indoor/outdoor, forward- and $45^\circ$-facing cameras); the event-camera and no-GT competition splits are omitted. GT-depth is stereo-derived.
Notably, \emph{TartanAirV2} (or its earlier version \emph{TartanAir}), appears in the original training corpus of all the backbones we use, except VGGT. Its inclusion aims to show that results, despite being closer to in-distribution rather than zero-shot, still exhibit a clear scene-specific gap we target. The datasets utilized in this work span the whole motion regime range: noise-free (synthetic) and high-parallax (\textit{TartanAirV2}); moderate motion (\textit{EuRoC}); and highly dynamic, low-parallax (\textit{UZH-FPV}).

\subsection{Experimental setup}
\label{sec:impl}


Consistent with the nature of 3DFMs, which operate on short context windows, we evaluate on \emph{clips}: short segments sampled from each sequence consisting of a window of $n{=}8$ frames, with a stride of two. We sample three clips per sequence (near the start, middle, and end) for diverse coverage. All reported metrics are therefore aggregated as the median over all clips across a dataset's sequences.

\noindent\textbf{Backbones.} We extensively analyze the behaviour of test-time refinement on five relevant pretrained 3DFMs (using their released weights): three of them up-to-scale, VGGT~\cite{wang_vggt_2025}, VGGT-$\Omega$~\cite{wang_vggt-_2026} and $\pi^3$~\cite{wang_pi3_2025}; and other two scale-aware, Pi3X~\cite{wang_pi3_2025} and Depth Anything 3 (DA3)~\cite{lin_depthA3_2025}. 

\noindent\textbf{Optimization.} For the TTR, we experimentally set $w_r{=}20$, $w_t{=}10$, $w_{\text{d}}{=}10$, $w_{a_p}{=}1000$, and $w_{a_d}{=}1$, as they bring a good trade-off between pose refinement and scale recovery, without corrupting depth. We optimize with AdamW for a fixed budget of $K=160$ steps per clip. Experiments run on a single NVIDIA RTX~5090. TTR optimizes the scale factor $\alpha$, alongside the model’s camera head, depth head, and the final aggregator blocks ($L = 8$), while keeping all other parameters frozen.


\noindent\textbf{Evaluation metrics.} We evaluate camera pose, scale, and dense depth. Unlike conventional visual-odometry practice, we do not align trajectories before computing pose error, as our goal is to assess absolute metric accuracy. \emph{Trans} is the RMSE of per-frame camera positions against GT (meters); \emph{Rot} is the mean geodesic angle to GT orientations (degrees). \emph{Scale (Traj.)} is the median per-frame ratio $|t_\text{pred}|/|t_\text{gt}|$, where $1$ is ideal and values below/above indicate under/over-scaling, excluding near-stationary (GT displacement ${<}1$cm) clips, for which the ratio is undefined. For dense depth we report two complementary metrics. Shape is invariant to global scale and isolates geometric fidelity, measured via per-frame median scale-alignment \emph{AbsRel} ($\downarrow$) and the absolute inlier ratio \emph{$\delta{<}1.25$} ($\uparrow$), which measures the pixels within $1.25\times$ of GT without alignment. \emph{Scale (Depth)}, measured without alignment as $\operatorname{median}(D_{\text{pred}}/D_{\text{gt}})$, again targets $1$. Reporting them separately distinguishes a reconstruction correctly shaped but mis-scaled, from one correct on both, only differentiated by the \emph{Scale (Depth)} column.

All results are seeded; on fixed hardware, re-runs report variances within 0.01 m, 0.04$^\circ$  and 0.03 in scale, far below the reported anchoring effects.


\subsection{Results}
\label{sec:benchmark}

\begin{figure}
    \centering
    \includegraphics[width=0.4\textwidth]{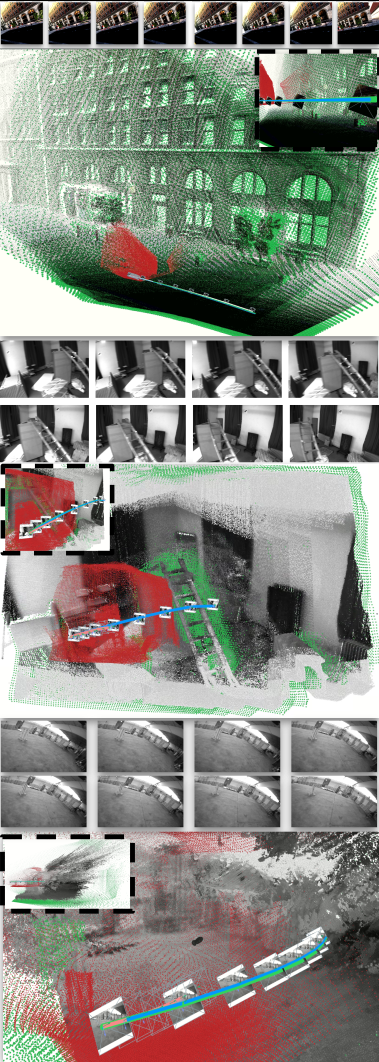}
    \caption{Qualitative examples. From top to bottom: TartanAirV2 ModernCityDowntown P000; EuRoC Vicon-Room V2-03; UZH-FPV indoor-forward-9. VI3 (green) correctly recovers, thanks to the inertial signal (blue), metric, scene-consistent geometry, in contrast to the pretrained 3DFM (red). Best in color.}
    \label{fig:qualitative_results}
    \vspace{-8mm}
\end{figure}

Fig.~\ref{fig:qualitative_results} illustrates qualitative results, further examples can be found in the supplementary. 
Tables~\ref{tab:non-metric} and~\ref{tab:metric} report the median results respectively for up-to-scale and metric backbones across sequences and clips of the raw pretrained model (\emph{Base}), IMU conditioning (if applicable) and our TTR. The results show that inertial measures are useful to inject physical knowledge in models spanning both natures. 
\emph{Up-to-scale} models can only recover a correct geometry, but at most up to a 45\% of the original scale, reflected also by position and the poor results in  $\delta{<}1.25$, which confirms the monocular scale ambiguity. In turn, our proposed refinement is able to successfully retrieve near metric reconstructions in all cases (scale $\approx 1$), which is propagated to translation. Additionally, we observe a consistent improvement in rotation, while our proposed cross-view loss helps to preserve the geometry, meaning that the injected scale does not corrupt 3D awareness. 
\emph{Metric-aware} models produce scale estimates closer to 1 and much better translation results (especially on \textit{TartanAirV2}, consistent with its in-distribution status), but we show that the IMU can improve this substantially in out-of-distribution cases both by TTR or conditioning on auxiliary geometry at test time. We also observe a slight but consistent improvement in the rotation, meaning that the inertial measurements can inject accurate information for the six degrees of freedom of the motion. We can see that, in some cases, the performance does not vary by just conditioning the network by metric inputs, suggesting that the data-driven prior is overriding any input. In these cases, TTR proves to be usually a better alternative, since the backbone parameters are adapted to fit the specific scene.

\subsection{Additional analyses}

\begin{figure*}[ht!]
    \centering
    \includegraphics[width=1.0\textwidth]{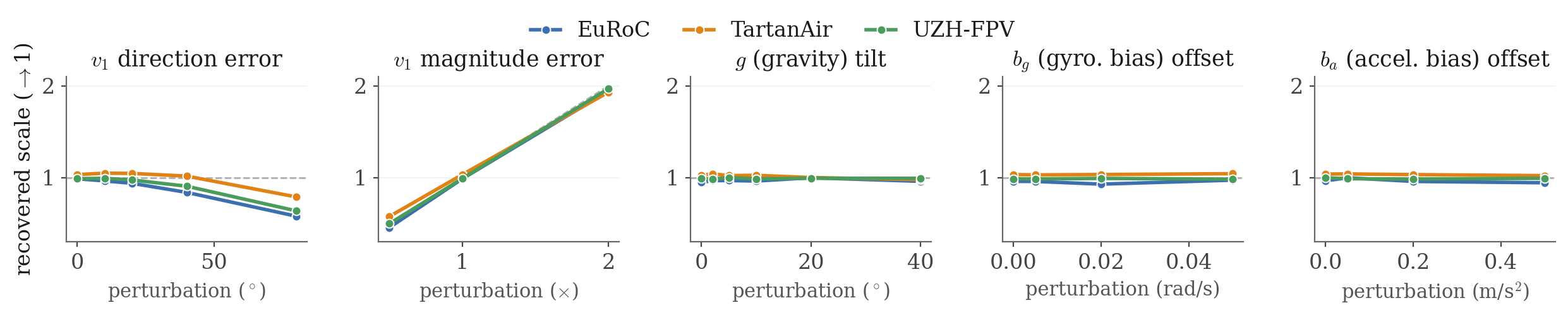}
    \caption{\textbf{Sensitivity to inertial initialization of recovered metric scale (VGGT backbone).} Downstream scale ($\rightarrow 1$ ideal) as each GT initialization component is independently perturbed (growing values of error applied to initial linear velocity, gravity and IMU biases).}
    \label{fig:init_sensitivity}
\end{figure*}

The most relevant analyses are included here, while further ones can be found in the supplementary material.





\noindent\textbf{Inertial initialization sensitivity.} We test how IMU initialization errors affect the recovered metric scale, applying varied noise over the preintegrated trajectory, then injected in a VGGT backbone using TTR. Fig.~\ref{fig:init_sensitivity} shows that VI3's scale recovery is robust against IMU biases and gravity noise, while being greatly affected by deviations in the initial velocity estimate: direction and especially magnitude errors degrade performance substantially.

\noindent\textbf{Initialization accuracy.} We evaluate the error obtained by the initialization procedure described in \cref{sec:init}, analyzing two sensor-derived variables, less dependent on the choice of backbone: absolute error angle in the gravity direction $\hat{\mathbf{g}}$ and the relative gyro bias error $\|\hat{b}_g-b_g\|/\|b_g\|$. Tab.~\ref{tab:imu_calib} shows the results (only \textit{EuRoC} has GT bias), where we observe that our initialization process achieves good estimates for both variables, with a worse performance in the \textit{UZH-FPV} dataset, more challenging due to the high speed.

We also evaluate the error obtained in the estimation of the most sensitive initialization parameter, the initial velocity. Since this metric depends on the employed backbone, we evaluate the error in both components: direction and magnitude factor ($||\hat v_0|-|v_0||/|v_0|$). \cref{tab:imu_init} shows that the direction error stays small across backbones while the magnitude spreads more spreads widely.

%% file: tables/metric.tex


\begin{table*}
\centering
\small
\caption{\textbf{Metric models: learned priors vs.\ inertial cues.} Each backbone's native metric prediction (\textbf{Base}), the same pretrained model conditioned on our GT-free IMU-integrated poses (\textbf{IMU-cond}, pose-only, feed-forward), and, conversely, using inertial test-time refinement (\textbf{TTR}). $^\dagger$IMU-cond additionally feeds camera intrinsics (required by DA3's pose encoder); Pi3X's IMU-cond is pose-only.}
\label{tab:metric}
\setlength{\tabcolsep}{4pt}
\resizebox{\textwidth}{!}{%
\begin{tabular}{lccccccccccccc}
\toprule
 \multirow{2}{*}{Dataset}  &  \multirow{2}{*}{Model}  & \multicolumn{3}{c}{Trans (m) $\downarrow$} & \multicolumn{3}{c}{Rot ($^\circ$) $\downarrow$} & \multicolumn{3}{c}{Scale (Traj.) $\rightarrow$1} & \multicolumn{3}{c}{Scale (Depth) $\rightarrow$1} \\
\cmidrule(lr){3-5}\cmidrule(lr){6-8}\cmidrule(lr){9-11}\cmidrule(lr){12-14}
 & & Base & IMU-cond & TTR & Base & IMU-cond & TTR & Base & IMU-cond & TTR & Base & IMU-cond & TTR \\
\midrule
 \multirow{ 2}{*}{\emph{TartanAirV2}} & \quad Pi3X & 0.392 & \textbf{0.371} & 0.382 & 0.802 & 0.546 & \textbf{0.316} & 1.18 & 1.20 & \textbf{1.08} & 1.26 & 1.25 & \textbf{1.11} \\
  & \quad DA3 & \textbf{0.322} & 0.446$^\dagger$ & 0.486 & 1.379 & 0.801$^\dagger$ & \textbf{0.351} & 0.90 & 0.83$^\dagger$ & \textbf{1.06} & \textbf{0.99} & 0.90$^\dagger$ & 1.19 \\
\midrule
  \multirow{ 2}{*}{\emph{EuRoC}} & \quad Pi3X & 0.071 & 0.081 & \textbf{0.067} & 0.257 & 0.306 & \textbf{0.254} & 0.83 & 0.82 & \textbf{1.04} & 0.91 & \textbf{0.93} & 1.11 \\
  & \quad DA3 & 0.061 & \textbf{0.055}$^\dagger$ & 0.092 & \textbf{0.261} & 0.326$^\dagger$ & 0.312 & 0.72 & 0.75$^\dagger$ & \textbf{1.13} & 0.74 & 0.73$^\dagger$ & \textbf{1.20} \\
\midrule
  \multirow{ 2}{*}{\emph{UZH-FPV}} & \quad Pi3X & 0.576 & 0.535 & \textbf{0.368} & 0.933 & 0.738 & \textbf{0.233} & \textbf{1.11} & 1.15 & 0.82 & 1.63 & 1.65 & \textbf{1.10} \\
  & \quad DA3 & 0.538 & 0.452$^\dagger$ & \textbf{0.379} & 1.252 & 0.626$^\dagger$ & \textbf{0.231} & \textbf{0.96} & 0.83$^\dagger$ & 0.88 & 1.39 & \textbf{1.05}$^\dagger$ & 1.07 \\
\bottomrule
\end{tabular}}
\end{table*}

%% file: tables/calib.tex
\begin{table}
\centering
\small
\caption{\textbf{Calibration-parameter init error.} Accuracy of the backbone-independent, fully-GT-free estimates of the two IMU calibration parameters that seed refinement: gravity direction and gyro bias, the latter as absolute error with relative in parentheses. GT gyro bias is only available on EuRoC (-- elsewhere).}
\label{tab:imu_calib}
\setlength{\tabcolsep}{6pt}
\begin{tabular}{lcc}
\toprule
Dataset & Gravity ($^\circ$) $\downarrow$ & Gyro-bias (rad/s) $\downarrow$ \\
\midrule
\textit{TartanAirV2} & 1.05 & -- \\
\textit{EuRoC} & 1.08 & 0.009~(11.3\%) \\
\textit{UZH-FPV} & 6.61 & -- \\
\bottomrule
\end{tabular}
\end{table}

%% file: tables/init.tex
\begin{table}
\centering
\small
\caption{\textbf{Initial velocity estimation.} Recovery of the direction and magnitude, for each specific dataset and backbone.}
\label{tab:imu_init}
\setlength{\tabcolsep}{3pt}
\resizebox{\columnwidth}{!}{%
\begin{tabular}{llccccc}
\toprule
\multirow{2}{*}{Dataset} &  \multirow{2}{*}{Metric} & \multicolumn{3}{c}{Up-to-scale} & \multicolumn{2}{c}{Scale-aware} \\
\cmidrule(lr){3-5}\cmidrule(lr){6-7}
& & VGGT & VGGT-$\Omega$ & $\pi$³ & Pi3X & DA3 \\
\midrule
\multirow{2}{*}{\textit{TartanAirV2}} 
 & $v_1$ dir. ($^\circ$) $\downarrow$ & 6.2 & 4.3 & 5.4 & 6.6 & 11.0 \\
 & $v_1$ mag. $\downarrow$           & 0.21 & 0.22 & 0.20 & 0.20 & 0.22 \\
\midrule
\multirow{2}{*}{\textit{EuRoC}} 
 & $v_1$ dir. ($^\circ$) $\downarrow$ & 10.0 & 3.6 & 9.3 & 8.8 & 5.6 \\
 & $v_1$ mag. $\downarrow$           & 0.21 & 0.23 & 0.26 & 0.17 & 0.22 \\
\midrule
\multirow{2}{*}{\textit{UZH-FPV}} 
 & $v_1$ dir. ($^\circ$) $\downarrow$ & 10.4 & 4.9 & 4.6 & 7.2 & 6.3 \\
 & $v_1$ mag. $\downarrow$           & 0.24 & 0.17 & 0.27 & 0.17 & 0.16 \\
\bottomrule
\end{tabular}
}

\end{table}

%% file: sec/5_conclusion.tex
\section{Conclusion}
\label{sec:conclusion}

In this work, we have demonstrated that the IMU is a suitable source of physical knowledge to anchor scale-ambiguous geometric outputs from any pretrained 3DFM into metric observations. For that, we introduced VI3, a sensor-in-the-loop framework that reframes physical anchoring, with diverse ways of injecting metric scale, including a dedicated refinement method. Through extensive experiments, we show that VI3 generalizes across backbones of different nature and diverse scenes; and that neither learned scale priors nor input conditioning are reliable enough alternatives, a reminder that physical properties need to be anchored in physical measurement, as data-driven priors can be inaccurate and prone to out-of-distribution failures.

The same philosophy extends as future work to any metric signal available at deployment, or a fusion of several (e.g. GNSS, wheel odometry, depth), and to downstream pipelines (feedforward novel-view synthesis, mapping) that depend on accurate metric geometry, pointing toward foundation models that specialize to the physical world they are dropped into.

%% file: sec/X_suppl.tex
\clearpage
\setcounter{page}{1}
\maketitlesupplementary

\label{sec:X_suppl}



\section*{A. Additional results}

We provide further experiments and ablation studies, spanning the analysis of specific components of VI3, specifically the inertial initialization and the test-time refinement.

\subsection*{A.1. Inertial initialization}

\noindent\textbf{Ground-truth inertial initialization}

Tables~\ref{tab:gt_upscale} and \ref{tab:gt_scaleaware} compare VI3 using our ground-truth-free (GT-free) inertial initialization from \cref{sec:init} against using the ground truth (GT) inertial parameters. As before, we separately assess its impact on TTR for the up-to-scale models, and on both input conditioning and TTR for the metric baselines. These results validate our proposed GT-free initialization, which maintains a small performance gap with respect to the GT oracle in rotation and depth, although the gap is larger for translation due to the sensitivity of the estimated scale to errors in the initial velocity. This trend is consistent across datasets and models, for both metric and up-to-scale settings. Overall, the results demonstrate that TTR can effectively absorb the errors introduced by the initial inertial estimate and refine the predictions, making it a more effective mechanism than input conditioning for injecting inertial cues into pretrained 3DFMs.

\noindent\textbf{Gravity consensus}

Table~\ref{tab:abl_guard} provides additional results for our gravity consensus mechanism. The image-based gravity estimation $\mathbf{g}_{\mathrm{gc}}$ typically fails on texture-less scenarios, as reflected by large mean errors in TartanAirV2 and UZH-FPV. The use of an alternative, rough, inertial-only gravity estimation ($\mathbf{g}_{\mathrm{imu}}$) brings the possibility to detect large errors of $\mathbf{g}_{\mathrm{gc}}$, and swap to a visual-inertial least-squares estimate ($\mathbf{g}_{\mathrm{ls}}$), which prevents from using largely erroneous values and, on average, shows a configuration which outperforms every gravity estimate alone. Note, in the right part of the table, the specific percentage of gravity swaps from $\mathbf{g}_{\mathrm{gc}}$ to $\mathbf{g}_{\mathrm{ls}}$, and how the error is substantially reduced for these specific cases. This shows how our consensus method effectively detects large errors from image-based gravity preditions.

\noindent\textbf{Initial velocity-scale estimation}

Table. \ref{tab:abl_ols_tls} supports our decision of formulating the initial velocity estimation as total least squares (TLS), instead of as an ordinary least squares (OLS). The assumption underlying TLS is that both the visual and inertial counterparts are noisy. On TartanAirV2, for which IMU data is noiseless, this assumption does not hold and both scale estimates show similar errors. However, on the two real (noisy) datasets (EuRoC and UZH-FPV) TLS leads to clear improvements over OLS, with scale estimates closer to the ideal value (1).

\noindent\textbf{Initial velocity magnitude}

When applying the regressed velocity scale to obtain the magnitude of the initial velocity, two options arise: scaling only the first pair of frames, or backprojecting all pairs to the initial one and taking the median as a robust representative. Table~\ref{tab:abl_v0mag} shows that using all pairs brings slightly, but clearly lower translation errors and a better scale estimates for out-of-distribution scenes with moderate parallax, in particular in EuRoC.

\input{tables/suppl_init_params_ablations}

\input{tables/suppl_gt_init_ablations}

\subsection*{A.2. Test-time refinement}

\noindent\textbf{Supervision objective}

Table~\ref{tab:loss_ablation} breaks down the supervision objective adopted in the our TTR. Exclusively supervising pose (keeping depth frozen) limits the benefits that multiple views can bring into the refinement process. Adding reprojection depth consistency, unlocks multi-view constraints in a self-supervised manner, leading to pose improvement. However, the inclusion of this up-to-scale loss degrades the scale estimates. Our adopted configuration solves this problem by regularizing depth, leading to a significant growth in depth metric correctness ($\delta{<}1.25$).




\noindent\textbf{Frame stride: parallax}

\begin{figure*}[ht!]
    \centering
    \includegraphics[width=1.0\textwidth]{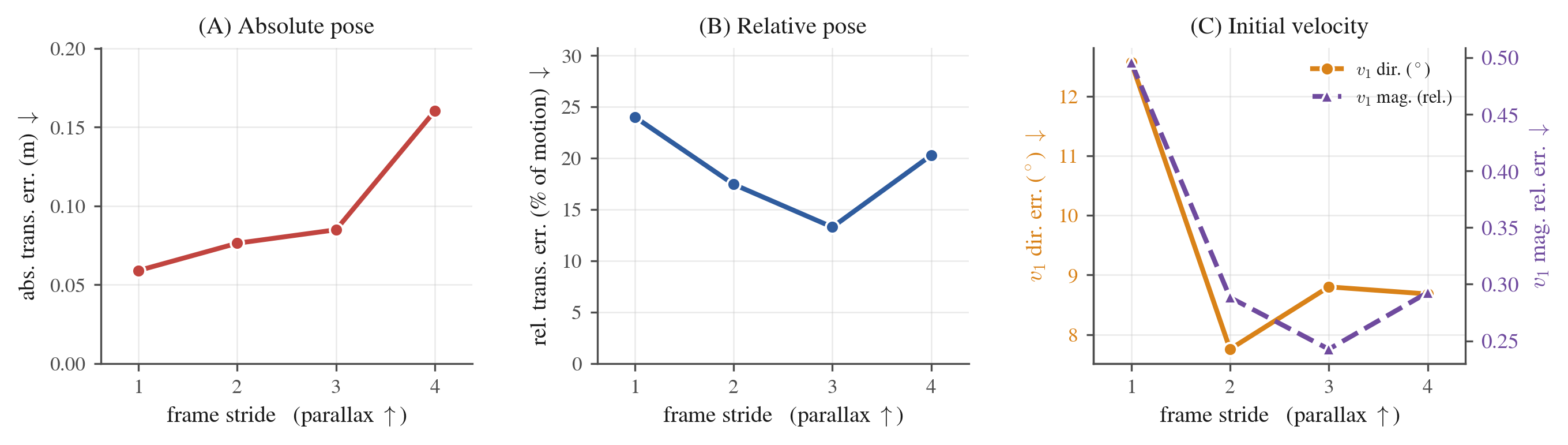}
    \caption{Frame-stride (parallax) effect absolute and relative poses, and initial velocity estimation.}
    \label{fig:stride}
\end{figure*}

Figure~\ref{fig:stride} analyzes the effect of the frame stride parameter, a direct tuning knob which allows adjusting parallax by skipping a number of consecutive sequence frames. {Figure~\ref{fig:stride}~(A)} shows absolute pose error increases with growing stride, which is a consequence of the absolute translation being bigger. Figure~\ref{fig:stride}~(B) normalizes the error by the translation length, which makes a more relevant pattern to emerge. Strides of 3 or 2 bring a good balance between parallax and translation in the datasets considered, as below them the parallax is insufficient, and above it the multi-view overlap diminishes and perspective effects change texture. The initial velocity, influenced by the translation accuracy commented, shows the same pattern, with stride 2 being optimal for initial velocity direction, and 3 for its magnitude counterpart, as Figure~\ref{fig:stride}~(C) exposes. We set stride 2 in the results presented in the main paper.

\noindent\textbf{Computational cost}

On Table~\ref{tab:timing} we show the backbone-specific initialization and test-time refinement cost. We use the consumer-grade NVIDIA RTX 5090 GPU. Note some models ($\dagger$) implement gradient checkpointing to fit in memory, increasing the cost for the choice of consumer-grade hardware. We also show, for the lightest model, VGGT-$\Omega$, the impact of fine-tuning capacity. More unfrozen blocks (higher capacity), implies longer overall processing time and a higher memory consumption. We find 8 to be a sweet spot between refinement capacity (see rotation column) but also computational time and memory, as for the choice of $n=8$ context frames, 8 is the upper bound to fit in memory some of the models utilized in this work.

\input{tables/suppl_ttr_ablations}

\section*{B. Additional qualitative results}

\begin{figure*}[ht!]
    \centering
    \includegraphics[width=1.0\textwidth]{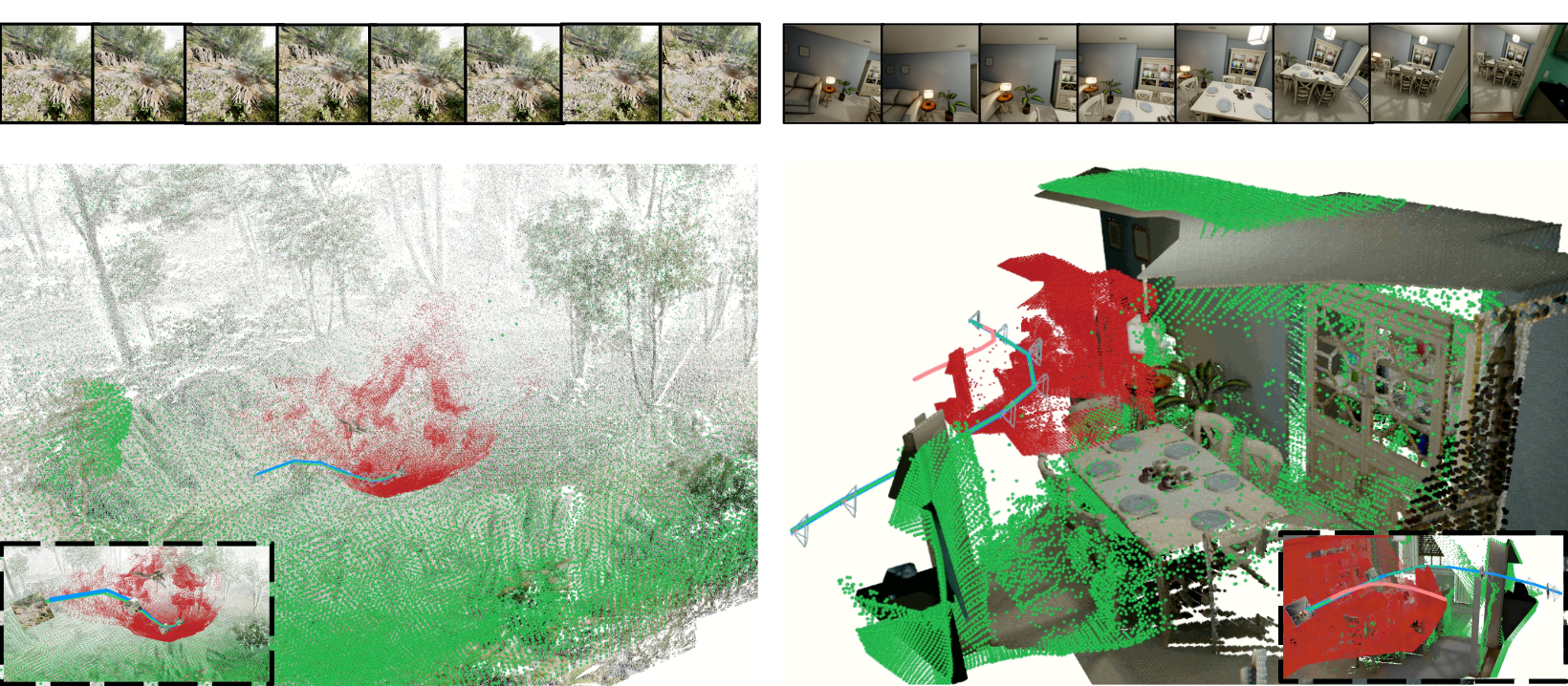}
    \caption{TartanAirV2 qualitative results: \textit{ForestEnv} and \textit{House} environments. Pretrained 3DFM (red); inertial trajectory (blue); VI3 (green); GT pointcloud (normal texture).}
    \label{fig:qualitative_tartanair_suppl}
\end{figure*}

\begin{figure*}[ht!]
    \centering
    \includegraphics[width=1.0\textwidth]{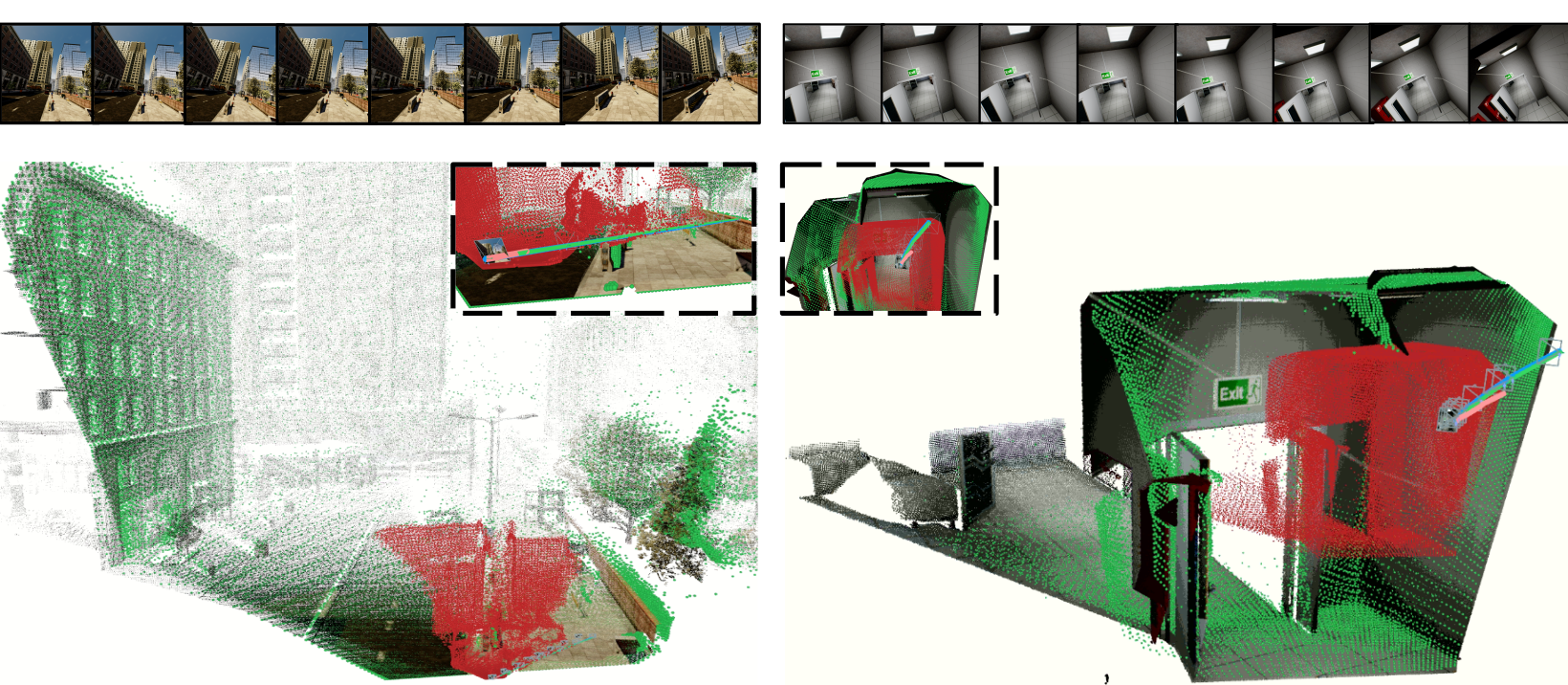}
    \caption{TartanAirV2 qualitative results: \textit{ModernCityDowntown} and \textit{Hospital} environments. Pretrained 3DFM (red); inertial trajectory (blue); VI3 (green); GT pointcloud (normal texture).}
    \label{fig:qualitative_tartanair_suppl2}
\end{figure*}

\begin{figure*}[ht!]
    \centering
    \includegraphics[width=1.0\textwidth]{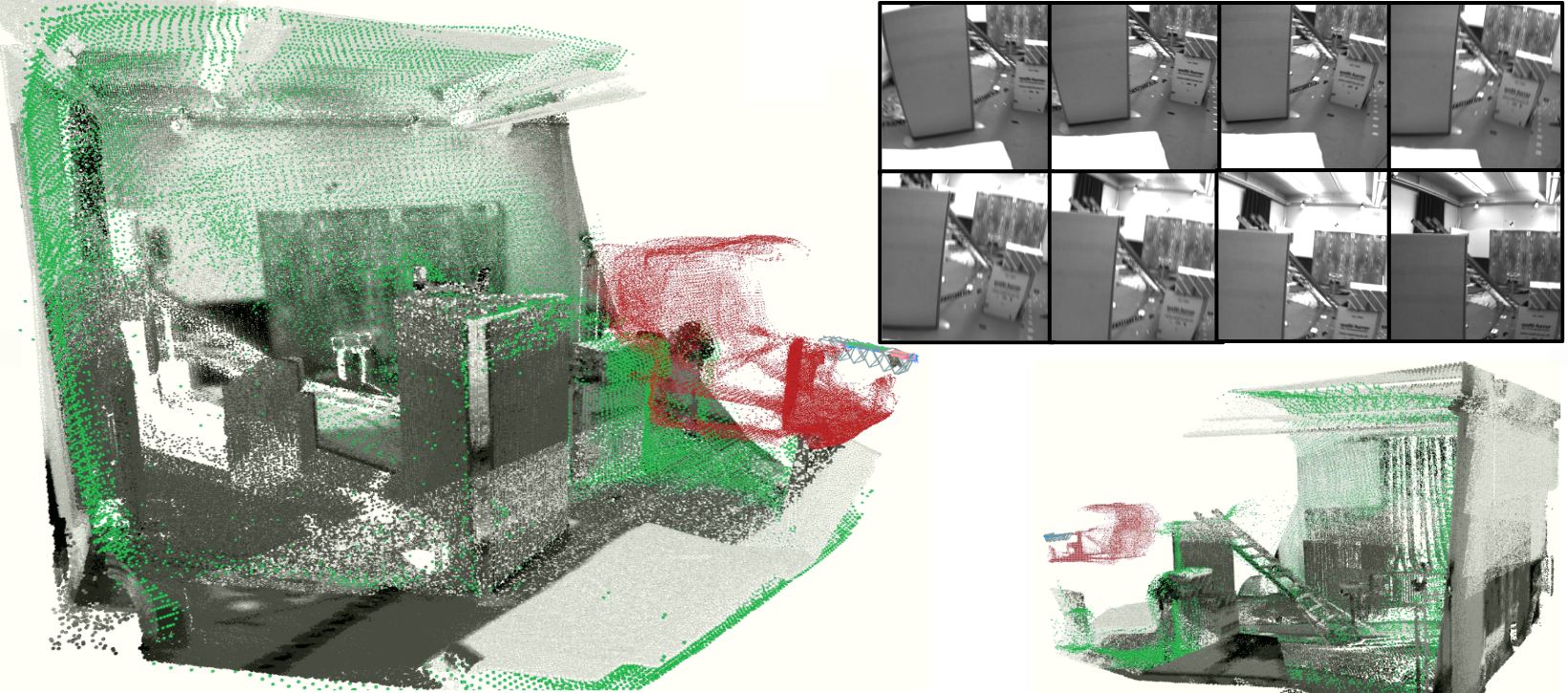}
    \caption{EuRoC qualitative results: \textit{Vicon Room 2} scene. Pretrained 3DFM (red); inertial trajectory (blue); VI3 (green); GT pointcloud (normal texture).}
    \label{fig:qualitative_euroc_suppl}
\end{figure*}

\begin{figure*}[ht!]
    \centering
    \includegraphics[width=1.0\textwidth]{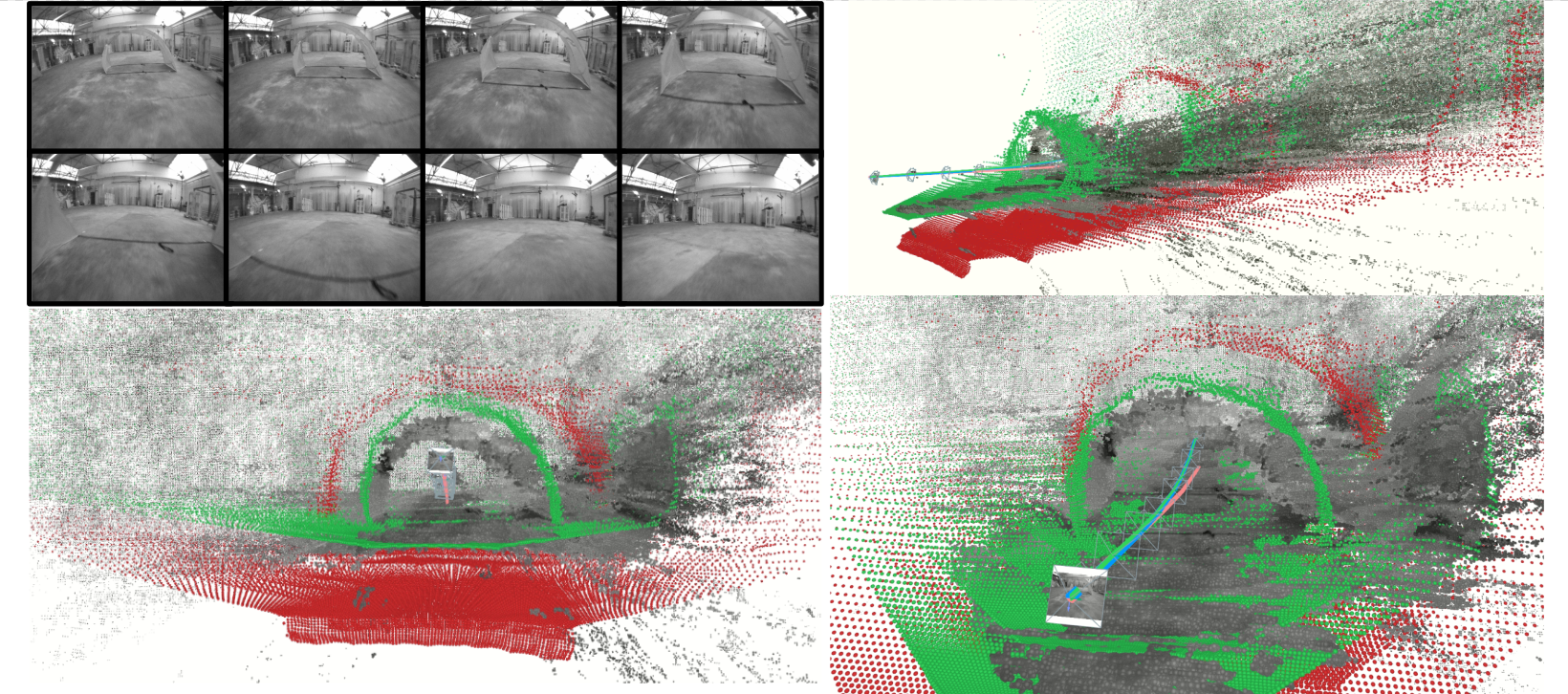}
    \caption{UZH-FPV qualitative results: \textit{Indoor Forward} scene. Pretrained 3DFM (red); inertial trajectory (blue); VI3 (green); GT pointcloud (normal texture), stereo-derived (noisy).}
    \label{fig:qualitative_uzh_suppl}
\end{figure*}

We show further examples of qualitative results. Figures~\ref{fig:qualitative_tartanair_suppl} and ~\ref{fig:qualitative_tartanair_suppl2} correspond to the four synthetic environments we evaluate on from TartanAirV2 (\textit{ForestEnv}, \textit{ModernCityDowntown}, \textit{Hospital} and \textit{House}).
Figure~\ref{fig:qualitative_euroc_suppl} presents an additional \textit{Vicon Room 2} reconstruction, with moderate motion and various scene objects; and Figure~\ref{fig:qualitative_uzh_suppl} shows how VI3 is capable of recovering from an overscaled metric-aware pretrained baseline, despite the high-speed motion profile of UZH-FPV \textit{Indoor Forward} scenes.


\section*{C. Limitations}

Our approach materializes a way to add inertial cues to pretrained 3DFMs without any ground-truth. We have shown VI3 robustly recovers metric motion and geometry across backbones and datasets. It is, however, bounded by what visual-inertial cues cannot observe. Metric scale is only recoverable where motion provides sufficient translational parallax; a fundamental observability limit rather than a failure of estimation. Models differ in how gracefully they degrade in this regime, since the estimate operates on their predicted inter-frame velocities.

%% file: tables/suppl_init_params_ablations.tex

\begin{table}[htbp!]
\centering
\small
\caption{\textbf{Init: gravity consensus.} VGGT model. Mean frame-0 gravity error ($^\circ$ vs.\ GT). For diverging consensus, we report \textbf{Fires} on \% of windows, replacing the default single-image \textbf{$\mathbf{g}_{\mathrm{gc}}$} with the motion-compensated \textbf{$\mathbf{g}_{\mathrm{ls}}$} only there; our estimate beats both sources over the full suite. Mean over sequences$\times$windows.}
\label{tab:abl_guard}
\setlength{\tabcolsep}{2pt}
\begin{tabular}{l cccc @{\hskip 8pt} cc}
\toprule
& \multicolumn{4}{c}{Full-suite ($^\circ$) $\downarrow$} & \multicolumn{2}{c}{Diverging consensus} \\
\cmidrule(lr){2-5}\cmidrule(lr){6-7}
Dataset & $\mathbf{g}_{\mathrm{gc}}$ & $\mathbf{g}_{\mathrm{imu}}$ & $\mathbf{g}_{\mathrm{ls}}$ & Ours & Fires (\%) & $\mathbf{g}_{\mathrm{gc}}$$\to$$\mathbf{g}_{\mathrm{ls}}$ ($^\circ$)\\
\midrule
TartanAirV2 & 9.67 & 8.58 & 4.31 & \textbf{1.96} & 10 & 78.3$\to$\textbf{4.3} \\
EuRoC & 1.98 & 4.22 & 2.62 & \textbf{1.98} & 0 & -- \\
UZH-FPV & 23.75 & 22.38 & 15.52 & \textbf{10.32} & 31 & 61.8$\to$\textbf{18.8} \\
\bottomrule
\end{tabular}
\end{table}


\begin{table}[htbp!]
\centering
\small
\caption{\textbf{Init: velocity-scale estimation -- OLS vs.\ TLS.} VGGT model. Implied $|\hat v_1|/|v_1^{\text{gt}}|$ (ideal $1.0$) of the ordinary- and total-least-squares scalar scale. OLS attenuates under predictor noise (regression dilution) $\Rightarrow$ systematic under-scale; the errors-in-variables TLS is unbiased.}
\label{tab:abl_ols_tls}
\begin{tabular}{lcc}
\toprule
Dataset & OLS & TLS \\
\midrule
TartanAirV2 & \textbf{0.89} & 1.12 \\
EuRoC & 0.76 & \textbf{0.96} \\
UZH-FPV & 0.60 & \textbf{0.70} \\
\bottomrule
\end{tabular}
\end{table}


\begin{table}[htbp!]
\centering
\small
\caption{\textbf{Init: velocity magnitude.} Using the TLS scale on the first pair alone (\textbf{first-pair}, Eq.~\ref{eq:S}) vs.\ the robust median back-projection over all pairs (\textbf{all-pairs}, Eq.~\ref{eq:v0}. The all-pairs back-out lowers translation error and pulls both trajectory and depth scale toward $1$ where the single-pair estimate is noisy. VGGT.}
\label{tab:abl_v0mag}
\setlength{\tabcolsep}{4pt}
\begin{tabular}{l cc cc cc}
\toprule
& \multicolumn{2}{c}{Trans (m) $\downarrow$} & \multicolumn{2}{c}{\footnotesize Scale ({\scriptsize Traj.}) $\rightarrow$1} & \multicolumn{2}{c}{\footnotesize Scale ({\scriptsize Depth}) $\rightarrow$1} \\
\cmidrule(lr){2-3}\cmidrule(lr){4-5}\cmidrule(lr){6-7}
Dataset & first & all & first & all & first & all \\
\midrule
TartanAirV2 & 0.505 & \textbf{0.351} & \textbf{1.00} & 1.05 & \textbf{1.05} & 1.11 \\
EuRoC & 0.060 & 0.060 & 0.91 & \textbf{0.99} & 1.08 & \textbf{1.00} \\
UZH-FPV & 0.537 & \textbf{0.495} & \textbf{0.81} & 0.80 & \textbf{0.94} & \textbf{0.94} \\
\bottomrule
\end{tabular}
\end{table}

%% file: tables/suppl_gt_init_ablations.tex

\begin{table*}[t]
\centering
\normalsize
\caption{\textbf{Initialization of up-to-scale models: GT-free vs.\ GT (Oracle).} Inertial test-time refinement with fully-GT-free init vs.\ an oracle GT init (GT gravity, $v_1$, gyro bias). Each cell is \emph{GT-free / GT}; better in bold. GT-free reaches near-oracle quality.}
\label{tab:gt_upscale}
\setlength{\tabcolsep}{6pt}
\begin{tabular}{@{}c l cccccc@{}}
\toprule
& Model & Trans (m) $\downarrow$ & Rot ($^\circ$) $\downarrow$ & \footnotesize Scale {\scriptsize(Traj.)} $\rightarrow$1 & AbsRel $\downarrow$ & $\delta{<}1.25$ $\uparrow$ & \footnotesize Scale ({\scriptsize Depth}) $\rightarrow$1 \\
\cmidrule(lr){3-8}
& Init & \footnotesize GT-free / GT & \footnotesize GT-free / GT & \footnotesize GT-free / GT & \footnotesize GT-free / GT & \footnotesize GT-free / GT & \footnotesize GT-free / GT \\
\midrule
\multirow{3}{*}{\rotatebox[origin=c]{90}{\footnotesize\emph{TartanAir}}} & VGGT & 0.358 / \textbf{0.058} & 0.410 / \textbf{0.348} & 1.05 / \textbf{0.99} & 0.089 / \textbf{0.076} & 0.515 / \textbf{0.882} & 1.11 / \textbf{1.06} \\
 & VGGT-$\Omega$ & 0.378 / \textbf{0.053} & 0.354 / \textbf{0.248} & 1.10 / \textbf{1.00} & 0.080 / \textbf{0.077} & 0.596 / \textbf{0.938} & 1.12 / \textbf{1.04} \\
 & $\pi$³ & 0.350 / \textbf{0.053} & \textbf{0.296} / 0.307 & 1.09 / \textbf{0.99} & 0.083 / \textbf{0.071} & 0.556 / \textbf{0.921} & 1.12 / \textbf{1.01} \\
\midrule
\multirow{3}{*}{\rotatebox[origin=c]{90}{\footnotesize\emph{EuRoC}}} & VGGT & 0.061 / \textbf{0.015} & 0.285 / \textbf{0.118} & 0.98 / \textbf{1.00} & \textbf{0.234} / 0.240 & 0.324 / \textbf{0.465} & \textbf{1.00} / 1.08 \\
 & VGGT-$\Omega$ & 0.074 / \textbf{0.016} & 0.252 / \textbf{0.062} & 1.05 / \textbf{1.00} & \textbf{0.215} / 0.219 & 0.243 / \textbf{0.653} & 1.10 / \textbf{1.03} \\
 & $\pi$³ & 0.078 / \textbf{0.012} & 0.164 / \textbf{0.105} & 1.08 / \textbf{0.99} & \textbf{0.203} / 0.204 & 0.251 / \textbf{0.672} & 1.18 / \textbf{1.07} \\
\midrule
\multirow{3}{*}{\rotatebox[origin=c]{90}{\footnotesize\emph{UZH-FPV}}} & VGGT & 0.497 / \textbf{0.027} & \textbf{0.301} / 0.344 & 0.80 / \textbf{1.00} & \textbf{0.203} / 0.209 & 0.201 / \textbf{0.379} & \textbf{0.92} / 1.28 \\
 & VGGT-$\Omega$ & 0.328 / \textbf{0.009} & 0.192 / \textbf{0.189} & 0.92 / \textbf{1.00} & \textbf{0.237} / 0.242 & 0.217 / \textbf{0.309} & \textbf{1.20} / 1.30 \\
 & $\pi$³ & 0.396 / \textbf{0.022} & 0.246 / \textbf{0.242} & 0.79 / \textbf{1.00} & 0.249 / \textbf{0.239} & 0.138 / \textbf{0.247} & \textbf{1.07} / 1.34 \\
\bottomrule
\end{tabular}
\end{table*}


\begin{table*}[t]
\centering
\small
\caption{\textbf{Initialization of scale-aware models: GT-free vs.\ GT} IMU-pose conditioning (\textbf{IMU-cond}, feed-forward) and inertial test-time refinement (\textbf{TTR}), each shown as \emph{GT-free\,/\,GT}: our fully-GT-free init vs.\ an oracle init (GT gravity, $v_0$, gyro bias). Better in bold. $^\dagger$DA3's IMU-cond additionally feeds camera intrinsics (required by its pose encoder); Pi3X's is pose-only. Median over sequences$\times$windows.}
\label{tab:gt_scaleaware}
\setlength{\tabcolsep}{4pt}
\resizebox{\textwidth}{!}{%
\begin{tabular}{llcccccccc}
\toprule
\multirow{3}{*}{Dataset} & \multirow{3}{*}{Model} & \multicolumn{2}{c}{Trans (m) $\downarrow$} & \multicolumn{2}{c}{Rot ($^\circ$) $\downarrow$} & \multicolumn{2}{c}{Scale (Traj.) $\rightarrow$1} & \multicolumn{2}{c}{Scale (Depth) $\rightarrow$1} \\
\cmidrule(lr){3-4}\cmidrule(lr){5-6}\cmidrule(lr){7-8}\cmidrule(lr){9-10}
 & & IMU-cond & TTR & IMU-cond & TTR & IMU-cond & TTR & IMU-cond & TTR \\
 & & \footnotesize GT-free/GT & \footnotesize GT-free/GT & \footnotesize GT-free/GT & \footnotesize GT-free/GT & \footnotesize GT-free/GT & \footnotesize GT-free/GT & \footnotesize GT-free/GT & \footnotesize GT-free/GT \\
\midrule
\multirow{2}{*}{\emph{TartanAir}} & \quad Pi3X & 0.371 / \textbf{0.339} & 0.417 / \textbf{0.061} & \textbf{0.546} / 0.589 & \textbf{0.310} / 0.320 & \textbf{1.20} / 1.20 & 1.09 / \textbf{0.99} & \textbf{1.25} / 1.26 & 1.10 / \textbf{1.03} \\
 & \quad DA3 & 0.446 / \textbf{0.248}$^\dagger$ & 0.486 / \textbf{0.061} & 0.801 / \textbf{0.683}$^\dagger$ & 0.351 / \textbf{0.345} & \textbf{0.83} / 0.81$^\dagger$ & 1.06 / \textbf{1.01} & \textbf{0.90} / 0.88$^\dagger$ & 1.19 / \textbf{1.16} \\
\midrule
\multirow{2}{*}{\emph{EuRoC}} & \quad Pi3X & 0.081 / \textbf{0.070} & 0.071 / \textbf{0.015} & 0.306 / \textbf{0.257} & 0.247 / \textbf{0.138} & 0.82 / \textbf{0.85} & 1.02 / \textbf{1.00} & \textbf{0.93} / 0.93 & 1.12 / \textbf{1.07} \\
 & \quad DA3 & \textbf{0.055} / 0.057$^\dagger$ & 0.092 / \textbf{0.027} & 0.326 / \textbf{0.235}$^\dagger$ & 0.312 / \textbf{0.141} & 0.75 / \textbf{0.76}$^\dagger$ & 1.13 / \textbf{1.01} & \textbf{0.73} / 0.73$^\dagger$ & 1.20 / \textbf{1.04} \\
\midrule
\multirow{2}{*}{\emph{UZH-FPV}} & \quad Pi3X & 0.535 / \textbf{0.494} & 0.404 / \textbf{0.036} & \textbf{0.738} / 0.812 & 0.231 / \textbf{0.214} & 1.15 / \textbf{1.15} & 0.82 / \textbf{1.00} & 1.65 / \textbf{1.64} & \textbf{1.09} / 1.38 \\
 & \quad DA3 & 0.452 / \textbf{0.377}$^\dagger$ & 0.379 / \textbf{0.020} & 0.626 / \textbf{0.423}$^\dagger$ & 0.231 / \textbf{0.228} & 0.83 / \textbf{0.84}$^\dagger$ & 0.88 / \textbf{0.99} & \textbf{1.05} / 1.06$^\dagger$ & \textbf{1.07} / 1.27 \\
\bottomrule
\end{tabular}}
\end{table*}

%% file: tables/suppl_ttr_ablations.tex

\begin{table*}[htbp!]
\centering
\small
\caption{\textbf{TTR: supervision objective ablation.} Results shown for VGGT, on EuRoC, using GT-free init (VI3). Adding our self-supervised depth-consistency term to the pose loss improves trajectory refinement and metric depth; anchoring then fixes the metric-depth drift the consistency term alone erodes.}
\label{tab:loss_ablation}
\begin{tabular}{l cccccc}
\toprule
Loss & Trans (m) $\downarrow$ & Rot ($^\circ$) $\downarrow$ & Scale (Traj.) $\rightarrow$1 & AbsRel $\downarrow$ & $\delta{<}1.25$ $\uparrow$ & Scale (Depth) $\rightarrow$1 \\
\midrule
Pose-only                         & 0.078 & 0.441 & 0.983 & 0.240 & 0.275 & 0.994 \\
\quad $+$ SSL dcons             & 0.063 & 0.334 & 0.981 & 0.242 & 0.246 & 0.917 \\
\quad\quad $+$ Anchoring             & \textbf{0.062} & \textbf{0.290} & \textbf{0.988} & \textbf{0.234} & \textbf{0.324} & \textbf{1.003} \\
\bottomrule
\end{tabular}
\end{table*}




\begin{table}[t]\centering
\caption{\textbf{Test-time refinement cost} on a single NVIDIA RTX 5090, ($n=8$, stride $2$). \emph{Init} is the one-time GT-free initialization, with model-weight loading excluded as it is amortized across windows. \emph{ms / iteration} is the amortized per-step wall-time. \emph{Total} is the time to reach the converged operating point of $160$ iterations, i.e.\ Init $+\,160\times$(ms / iteration). The initialization is essentially free ($\sim\!230$\,ms) and model independent; the cost is dominated by the refinement steps. \textsuperscript{\dag}Uses gradient checkpointing to be able to fit $n=8$, trading recompute for memory (reflected in the per-iteration cost).}
\label{tab:timing}
\setlength{\tabcolsep}{7pt}
\begin{tabular}{lccc}
\toprule
\multicolumn{4}{l}{\emph{Initialization and optimization steps} } \\
\toprule
Model & Init & ms / iteration & Total ($160$ it.) \\
\midrule
VGGT\textsuperscript{\dag}      & 232\,ms & 706 & 113\,s \\
VGGT-$\Omega$                   & 232\,ms & 302 &  48\,s \\
$\pi^3$                         & 232\,ms & 492 &  79\,s \\
Pi3X\textsuperscript{\dag}      & 232\,ms & 739 & 118\,s \\
DA3\textsuperscript{\dag}       & 232\,ms & 802 & 128\,s \\
\midrule
\midrule
\multicolumn{4}{l}{\emph{Number of trainable blocks} (VGGT-$\Omega$)} \\
\midrule
Blocks & Rot.\ ($^\circ$) & ms / iteration & Total ($160$ it.) \\
\quad $0$   & 0.315 & 255 & 41\,s \\
\quad $4$   & 0.291 & 274 & 44\,s \\
\quad $8$ & 0.228 & 302 & 48\,s \\
\quad $16$  & 0.206 & 381 & 61\,s \\
\quad $24$ (Full)  & 0.234 & 421 & 67\,s \\
\bottomrule
\end{tabular}
\end{table}


%% file: example.bib
@String(CVPR= {IEEE Conf. Comput. Vis. Pattern Recog.})

@String(ICCV= {Int. Conf. Comput. Vis.})

@String(ECCV= {Eur. Conf. Comput. Vis.})

@String(CVPR  = {CVPR})

@String(ICCV  = {ICCV})

@String(ECCV  = {ECCV})

@article{kaiser2016simultaneous,
  title={Simultaneous state initialization and gyroscope bias calibration in visual inertial aided navigation},
  author={Kaiser, Jacques and Martinelli, Agostino and Fontana, Flavio and Scaramuzza, Davide},
  journal={IEEE Robotics and Automation Letters},
  volume={2},
  number={1},
  pages={18--25},
  year={2016},
  publisher={IEEE}
}


%% file: zotero.bib
@String { THREEDV      = {{Proceedings of the International Conference on 3D Vision ({3DV})}} }

@article{euroc-mav,
  author  = {Burri, Michael and Nikolic, Janosch and Gohl, Pascal and Schneider, Thomas and Rehder, Joern and Omari, Sammy and Achtelik, Markus W and Siegwart, Roland},
  title   = {The EuRoC micro aerial vehicle datasets},
  year    = {2016},
  doi     = {10.1177/0278364915620033},
  URL     = {http://ijr.sagepub.com/content/early/2016/01/21/0278364915620033.abstract},
  eprint  = {http://ijr.sagepub.com/content/early/2016/01/21/0278364915620033.full.pdf+html},
  journal = {The International Journal of Robotics Research}
}

@inproceedings{sattler2017large,
  title={Are large-scale 3d models really necessary for accurate visual localization?},
  author={Sattler, Torsten and Torii, Akihiko and Sivic, Josef and Pollefeys, Marc and Taira, Hajime and Okutomi, Masatoshi and Pajdla, Tomas},
  booktitle={2017 IEEE Conference on Computer Vision and Pattern Recognition (CVPR)},
  pages={6175--6184},
  year={2017},
  organization={IEEE}
}

@article{panek2026guide,
  title={A Guide to Structureless Visual Localization: V. Panek et al.},
  author={Panek, Vojtech and Zhou, Qunjie and Ding, Yaqing and Agostinho, S{\'e}rgio and Kukelova, Zuzana and Sattler, Torsten and Leal-Taix{\'e}, Laura},
  journal={International Journal of Computer Vision},
  volume={134},
  number={6},
  pages={263},
  year={2026},
  publisher={Springer}
}

@inproceedings{cut3r,
    Author = {Qianqian Wang* and Yifei Zhang* and Aleksander Holynski and Alexei A. Efros and Angjoo Kanazawa},
    Title = {Continuous 3D Perception Model with Persistent State},
    Year = {2025},
    booktitle={CVPR},
}

@inproceedings{han2025vittt,
  title={ViT$^3$: Unlocking Test-Time Training in Vision},
  author={Han, Dongchen and Li, Yining and Li, Tianyu and Cao, Zixuan and Wang, Ziming and Song, Jun and Cheng, Yu and Zheng, Bo and Huang, Gao},
  booktitle={Proceedings of the IEEE/CVF Conference on Computer Vision and Pattern Recognition},
  year={2026}
}

@inproceedings{jin2026zipmap,
    title     = {{ZipMap}: Linear-Time Stateful 3D Reconstruction via Test-Time Training},
    author    = {Jin, Haian and Wu, Rundi and Zhang, Tianyuan and Gao, Ruiqi and Barron, Jonathan T. and Snavely, Noah and Holynski, Aleksander},
    booktitle = {Proceedings of the IEEE/CVF Conference on Computer Vision and Pattern Recognition},
    year      = {2026}
}

@article{zhang2026loger,
  title={LoGeR: Long-Context Geometric Reconstruction with Hybrid Memory},
  author={Zhang, Junyi and Herrmann, Charles and Hur, Junhwa and Sun, Chen and Yang, Ming-Hsuan and Cole, Forrester and Darrell, Trevor and Sun, Deqing},
  journal={arXiv preprint arXiv:2603.03269},
  year={2026}
}

@article{monodepth2,
  title     = {Digging into Self-Supervised Monocular Depth Prediction},
  author    = {Cl{\'{e}}ment Godard and
               Oisin {Mac Aodha} and
               Michael Firman and
               Gabriel J. Brostow},
  journal = {The International Conference on Computer Vision (ICCV)},
  month = {October},
  year = {2019}
}

@misc{leroy_master_2024,
	title = {Grounding {Image} {Matching} in {3D} with {MASt3R}},
	url = {http://arxiv.org/abs/2406.09756},
	doi = {10.48550/arXiv.2406.09756},
	urldate = {2025-08-08},
	publisher = {arXiv},
	author = {Leroy, Vincent and Cabon, Yohann and Revaud, Jérôme},
	month = jun,
	year = {2024},
	note = {arXiv:2406.09756 [cs]},
}

@misc{wang_vggt_2025,
	title = {{VGGT}: {Visual} {Geometry} {Grounded} {Transformer}},
	shorttitle = {{VGGT}},
	url = {http://arxiv.org/abs/2503.11651},
	doi = {10.48550/arXiv.2503.11651},
	urldate = {2025-08-08},
	publisher = {arXiv},
	author = {Wang, Jianyuan and Chen, Minghao and Karaev, Nikita and Vedaldi, Andrea and Rupprecht, Christian and Novotny, David},
	month = mar,
	year = {2025},
	note = {arXiv:2503.11651 [cs]},
}

@inproceedings{schoenberger2016sfm,
    author={Sch\"{o}nberger, Johannes Lutz and Frahm, Jan-Michael},
    title={{Structure-from-Motion Revisited}},
    booktitle={Conference on Computer Vision and Pattern Recognition (CVPR)},
    year={2016}
}

@misc{wang_dust3r_2023,
	title = {{DUSt3R}: {Geometric} {3D} {Vision} {Made} {Easy}},
	shorttitle = {{DUSt3R}},
	url = {http://arxiv.org/abs/2312.14132},
	doi = {10.48550/arXiv.2312.14132},
	urldate = {2025-11-07},
	publisher = {arXiv},
	author = {Wang, Shuzhe and Leroy, Vincent and Cabon, Yohann and Chidlovskii, Boris and Revaud, Jerome},
	year = {2023},
	note = {arXiv:2312.14132 [cs]},
}

@misc{oquab_dinov2_2024,
	title = {{DINOv2}: {Learning} {Robust} {Visual} {Features} without {Supervision}},
	shorttitle = {{DINOv2}},
	url = {http://arxiv.org/abs/2304.07193},
	doi = {10.48550/arXiv.2304.07193},
	urldate = {2025-11-08},
	author = {Oquab, Maxime and Darcet, Timothée and Moutakanni, Théo and Vo, Huy and Szafraniec, Marc and Khalidov, Vasil and Fernandez, Pierre and Haziza, Daniel and Massa, Francisco and El-Nouby, Alaaeldin and Assran, Mahmoud and Ballas, Nicolas and Galuba, Wojciech and Howes, Russell and Huang, Po-Yao and Li, Shang-Wen and Misra, Ishan and Rabbat, Michael and Sharma, Vasu and Synnaeve, Gabriel and Xu, Hu and Jegou, Hervé and Mairal, Julien and Labatut, Patrick and Joulin, Armand and Bojanowski, Piotr},
	month = feb,
	year = {2024},
	note = {arXiv:2304.07193 [cs]},
}

@misc{lin_depthA3_2025,
	title = {Depth {Anything} 3: {Recovering} the {Visual} {Space} from {Any} {Views}},
	shorttitle = {Depth {Anything} 3},
	url = {http://arxiv.org/abs/2511.10647},
	doi = {10.48550/arXiv.2511.10647},
	urldate = {2025-11-18},
	publisher = {arXiv},
	author = {Lin, Haotong and Chen, Sili and Liew, Junhao and Chen, Donny Y. and Li, Zhenyu and Shi, Guang and Feng, Jiashi and Kang, Bingyi},
	month = nov,
	year = {2025},
	note = {arXiv:2511.10647 [cs]},
}

@misc{keetha_mapanything_2026,
	title = {{MapAnything}: {Universal} {Feed}-{Forward} {Metric} {3D} {Reconstruction}},
	shorttitle = {{MapAnything}},
	url = {http://arxiv.org/abs/2509.13414},
	doi = {10.48550/arXiv.2509.13414},
	urldate = {2026-03-09},
	publisher = {arXiv},
	author = {Keetha, Nikhil and Müller, Norman and Schönberger, Johannes and Porzi, Lorenzo and Zhang, Yuchen and Fischer, Tobias and Knapitsch, Arno and Zauss, Duncan and Weber, Ethan and Antunes, Nelson and Luiten, Jonathon and Lopez-Antequera, Manuel and Bulò, Samuel Rota and Richardt, Christian and Ramanan, Deva and Scherer, Sebastian and Kontschieder, Peter},
	month = jan,
	year = {2026},
	note = {arXiv:2509.13414 [cs]},
}

@misc{wang_pi3_2025,
  title={$\pi^3$: Permutation-Equivariant Visual Geometry Learning},
  author={Wang, Yifan and Zhou, Jianjun and Zhu, Haoyi and Chang, Wenzheng and Zhou, Yang and Li, Zizun and Chen, Junyi and Pang, Jiangmiao and Shen, Chunhua and He, Tong},
  booktitle={International Conference on Learning Representations},
  volume={2026},
  pages={10481--10497},
  year={2026}
}

@inproceedings{karhade2026any4d,
  title={Any4d: Unified feed-forward metric 4d reconstruction},
  author={Karhade, Jay and Keetha, Nikhil and Zhang, Yuchen and Gupta, Tanisha and Sharma, Akash and Scherer, Sebastian and Ramanan, Deva},
  booktitle={Proceedings of the IEEE/CVF Conference on Computer Vision and Pattern Recognition},
  pages={14578--14589},
  year={2026}
}

@inproceedings{pan2024global,
  title={Global structure-from-motion revisited},
  author={Pan, Linfei and Bar{\'a}th, D{\'a}niel and Pollefeys, Marc and Sch{\"o}nberger, Johannes L},
  booktitle={European Conference on Computer Vision},
  pages={58--77},
  year={2024},
  organization={Springer}
}

@article{mur2017orb,
  title={Orb-slam2: An open-source slam system for monocular, stereo, and rgb-d cameras},
  author={Mur-Artal, Raul and Tard{\'o}s, Juan D},
  journal={IEEE transactions on robotics},
  volume={33},
  number={5},
  pages={1255--1262},
  year={2017},
  publisher={IEEE}
}

@inproceedings{keetha2024splatam,
  title={Splatam: Splat, track \& map 3d gaussians for dense rgb-d slam},
  author={Keetha, Nikhil and Karhade, Jay and Jatavallabhula, Krishna Murthy and Yang, Gengshan and Scherer, Sebastian and Ramanan, Deva and Luiten, Jonathon},
  booktitle={2024 IEEE/CVF Conference on Computer Vision and Pattern Recognition (CVPR)},
  pages={21357--21366},
  year={2024},
  organization={IEEE}
}

@inproceedings{peng2026omnivggt,
  title={Omnivggt: Omni-modality driven visual geometry grounded transformer},
  author={Peng, Haosong and Li, Hao and Dai, Yalun and Lan, Yushi and Luo, Yihang and Qi, Tianyu and Zhang, Zhengshen and Zhan, Yufeng and Zhang, Junfei and Xu, Wenchao and others},
  booktitle={Proceedings of the IEEE/CVF Conference on Computer Vision and Pattern Recognition},
  pages={36485--36497},
  year={2026}
}

@article{agarwal2011building,
  title={Building rome in a day},
  author={Agarwal, Sameer and Furukawa, Yasutaka and Snavely, Noah and Simon, Ian and Curless, Brian and Seitz, Steven M and Szeliski, Richard},
  journal={Communications of the ACM},
  volume={54},
  number={10},
  pages={105--112},
  year={2011},
  publisher={ACM New York, NY, USA}
}

@article{bao20253d,
  title={3d gaussian splatting: Survey, technologies, challenges, and opportunities},
  author={Bao, Yanqi and Ding, Tianyu and Huo, Jing and Liu, Yaoli and Li, Yuxin and Li, Wenbin and Gao, Yang and Luo, Jiebo},
  journal={IEEE Transactions on Circuits and Systems for Video Technology},
  volume={35},
  number={7},
  pages={6832--6852},
  year={2025},
  publisher={IEEE}
}

@article{wang2026nerfs,
  title={NeRFs in robotics: A survey},
  author={Wang, Guangming and Pan, Lei and Peng, Songyou and Liu, Shaohui and Xu, Chenfeng and Miao, Yanzi and Zhan, Wei and Tomizuka, Masayoshi and Pollefeys, Marc and Wang, Hesheng},
  journal={The International Journal of Robotics Research},
  volume={45},
  number={6},
  pages={858--892},
  year={2026},
  publisher={SAGE Publications Sage UK: London, England}
}

@inproceedings{wang2026amb3r,
  title={AMB3R: Accurate feed-forward metric-scale 3D reconstruction with backend},
  author={Wang, Hengyi and Agapito, Lourdes},
  booktitle={Proceedings of the IEEE/CVF Conference on Computer Vision and Pattern Recognition},
  pages={14612--14625},
  year={2026}
}

@article{jang_pow3r_2025,
	title = {{Pow3R}: {Empowering} {Unconstrained} {3D} {Reconstruction} with {Camera} and {Scene} {Priors}},
	language = {en},
	author = {Jang, Wonbong and Weinzaepfel, Philippe and Leroy, Vincent and Agapito, Lourdes and Revaud, Jerome},
	year = {2025},
}

@misc{chen_ttt3r_2026,
	title = {{TTT3R}: {3D} {Reconstruction} as {Test}-{Time} {Training}},
	shorttitle = {{TTT3R}},
	url = {http://arxiv.org/abs/2509.26645},
	doi = {10.48550/arXiv.2509.26645},
	urldate = {2026-04-15},
	publisher = {arXiv},
	author = {Chen, Xingyu and Chen, Yue and Xiu, Yuliang and Geiger, Andreas and Chen, Anpei},
	month = mar,
	year = {2026},
	note = {arXiv:2509.26645 [cs]},
}

@misc{wang_vggt-_2026,
      title={{VGGT-$\Omega$}}, 
      author={Jianyuan Wang and Minghao Chen and Shangzhan Zhang and Nikita Karaev and Johannes Schönberger and Patrick Labatut and Piotr Bojanowski and David Novotny and Andrea Vedaldi and Christian Rupprecht},
      year={2026},
      eprint={2605.15195},
      archivePrefix={arXiv},
      primaryClass={cs.CV},
      url={https://arxiv.org/abs/2605.15195}, 
}

@inproceedings{uzh-fpv,
	address = {Montreal, QC, Canada},
	title = {Are {We} {Ready} for {Autonomous} {Drone} {Racing}? {The} {UZH}-{FPV} {Drone} {Racing} {Dataset}},
	copyright = {https://ieeexplore.ieee.org/Xplorehelp/downloads/license-information/IEEE.html},
	isbn = {978-1-5386-6027-0},
	shorttitle = {Are {We} {Ready} for {Autonomous} {Drone} {Racing}?},
	url = {https://ieeexplore.ieee.org/document/8793887/},
	doi = {10.1109/ICRA.2019.8793887},
	language = {en},
	urldate = {2026-06-26},
	booktitle = {2019 {International} {Conference} on {Robotics} and {Automation} ({ICRA})},
	publisher = {IEEE},
	author = {Delmerico, Jeffrey and Cieslewski, Titus and Rebecq, Henri and Faessler, Matthias and Scaramuzza, Davide},
	month = may,
	year = {2019},
	pages = {6713--6719},
}

@misc{wang_tartanair_2020,
	title = {{TartanAir}: {A} {Dataset} to {Push} the {Limits} of {Visual} {SLAM}},
	shorttitle = {{TartanAir}},
	url = {http://arxiv.org/abs/2003.14338},
	doi = {10.48550/arXiv.2003.14338},
	urldate = {2026-06-26},
	publisher = {arXiv},
	author = {Wang, Wenshan and Zhu, Delong and Wang, Xiangwei and Hu, Yaoyu and Qiu, Yuheng and Wang, Chen and Hu, Yafei and Kapoor, Ashish and Scherer, Sebastian},
	month = aug,
	year = {2020},
	note = {arXiv:2003.14338 [cs.RO]},
}

@article{forster_-manifold_2017,
	title = {On-{Manifold} {Preintegration} for {Real}-{Time} {Visual}-{Inertial} {Odometry}},
	volume = {33},
	issn = {1552-3098, 1941-0468},
	url = {http://arxiv.org/abs/1512.02363},
	doi = {10.1109/TRO.2016.2597321},
	number = {1},
	urldate = {2026-06-26},
	journal = {IEEE Transactions on Robotics},
	author = {Forster, Christian and Carlone, Luca and Dellaert, Frank and Scaramuzza, Davide},
	month = feb,
	year = {2017},
	note = {arXiv:1512.02363 [cs.RO]},
	pages = {1--21},
}

@article{Mur_Artal_2015_orbslam,
   title={ORB-SLAM: A Versatile and Accurate Monocular SLAM System},
   volume={31},
   ISSN={1941-0468},
   url={http://dx.doi.org/10.1109/TRO.2015.2463671},
   DOI={10.1109/tro.2015.2463671},
   number={5},
   journal={IEEE Transactions on Robotics},
   publisher={Institute of Electrical and Electronics Engineers (IEEE)},
   author={Mur-Artal, Raul and Montiel, J. M. M. and Tardos, Juan D.},
   year={2015},
   month=Oct, pages={1147–1163} }

@article{campos_orb-slam3_2021,
	title = {{ORB}-{SLAM3}: {An} {Accurate} {Open}-{Source} {Library} for {Visual}, {Visual}-{Inertial} and {Multi}-{Map} {SLAM}},
	volume = {37},
	issn = {1552-3098, 1941-0468},
	shorttitle = {{ORB}-{SLAM3}},
	url = {http://arxiv.org/abs/2007.11898},
	doi = {10.1109/TRO.2021.3075644},
	number = {6},
	urldate = {2026-06-26},
	journal = {IEEE Transactions on Robotics},
	author = {Campos, Carlos and Elvira, Richard and Rodríguez, Juan J. Gómez and Montiel, José M. M. and Tardós, Juan D.},
	month = dec,
	year = {2021},
	note = {arXiv:2007.11898 [cs.RO]},
	pages = {1874--1890},
}

@Conference{Geneva2020ICRAopenvins,
  Title      = {{OpenVINS}: A Research Platform for Visual-Inertial Estimation},
  Author     = {Patrick Geneva and Kevin Eckenhoff and Woosik Lee and Yulin Yang and Guoquan Huang},
  Booktitle  = {Proc. of the IEEE International Conference on Robotics and Automation},
  Year       = {2020},
  Address    = {Paris, France},
  Url        = {\url{https://github.com/rpng/open_vins}}
}

@inproceedings{ZHU2026VIGS-SLAM,
      title={VIGS-SLAM: Visual Inertial Gaussian Splatting SLAM},
      author={Zhu, Zihan and Zhang, Wei and Li, Moyang and Haala, Norbert and Pollefeys, Marc and Barath, Daniel},
      booktitle={European Conference on Computer Vision (ECCV)},
      year={2026}
}

@misc{zhou2025mast3rfusion,
      title={MASt3R-Fusion: Integrating Feed-Forward Visual Model with IMU, GNSS for High-Functionality SLAM}, 
      author={Yuxuan Zhou and Xingxing Li and Shengyu Li and Zhuohao Yan and Chunxi Xia and Shaoquan Feng},
      year={2025},
      eprint={2509.20757},
      archivePrefix={arXiv},
      primaryClass={cs.RO},
      url={https://arxiv.org/abs/2509.20757}, 
}

@inproceedings{wang2025spann3r,
  title={3d reconstruction with spatial memory},
  author={Wang, Hengyi and Agapito, Lourdes},
  booktitle=THREEDV,
  pages={78--89},
  year={2025},
  organization={IEEE}
}

@inproceedings{cabon2025must3r,
  title={Must3r: Multi-view network for stereo 3d reconstruction},
  author={Cabon, Yohann and Stoffl, Lucas and Antsfeld, Leonid and Csurka, Gabriela and Chidlovskii, Boris and Revaud, Jerome and Leroy, Vincent},
  booktitle=CVPR,
  pages={1050--1060},
  year={2025}
}

@inproceedings{liu2025slam3r,
  title={Slam3r: Real-time dense scene reconstruction from monocular rgb videos},
  author={Liu, Yuzheng and Dong, Siyan and Wang, Shuzhe and Yin, Yingda and Yang, Yanchao and Fan, Qingnan and Chen, Baoquan},
  booktitle=CVPR,
  pages={16651--16662},
  year={2025}
}

@article{zhuo2025streamvggt,
  title={Streaming 4d visual geometry transformer},
  author={Zhuo, Dong and Zheng, Wenzhao and Guo, Jiahe and Wu, Yuqi and Zhou, Jie and Lu, Jiwen},
  journal={arXiv preprint arXiv:2507.11539},
  year={2025}
}

@article{chen2025long3r,
  title={Long3r: Long sequence streaming 3d reconstruction},
  author={Chen, Zhuoguang and Qin, Minghui and Yuan, Tianyuan and Liu, Zhe and Zhao, Hang},
  journal={arXiv preprint arXiv:2507.18255},
  year={2025}
}

@article{lan2025stream3r,
  title={STream3R: Scalable Sequential 3D Reconstruction with Causal Transformer},
  author={Lan, Yushi and Luo, Yihang and Hong, Fangzhou and Zhou, Shangchen and Chen, Honghua and Lyu, Zhaoyang and Yang, Shuai and Dai, Bo and Loy, Chen Change and Pan, Xingang},
  journal={arXiv preprint arXiv:2508.10893},
  year={2025}
}

@article{li2025wint3r,
  title={WinT3R: Window-Based Streaming Reconstruction with Camera Token Pool},
  author={Li, Zizun and Zhou, Jianjun and Wang, Yifan and Guo, Haoyu and Chang, Wenzheng and Zhou, Yang and Zhu, Haoyi and Chen, Junyi and Shen, Chunhua and He, Tong},
  journal={arXiv preprint arXiv:2509.05296},
  year={2025}
}

@article{wu2025point3r,
  title={Point3R: Streaming 3D Reconstruction with Explicit Spatial Pointer Memory},
  author={Wu, Yuqi and Zheng, Wenzhao and Zhou, Jie and Lu, Jiwen},
  journal={arXiv preprint arXiv:2507.02863},
  year={2025}
}

@article{ma2025puzzles,
  title={Puzzles: Unbounded Video-Depth Augmentation for Scalable End-to-End 3D Reconstruction},
  author={Ma, Jiahao and Wang, Lei and Ahmedt-Aristizabal, David and Nguyen, Chuong and others},
  journal={arXiv preprint arXiv:2506.23863},
  year={2025}
}

@article{khafizov2025g,
  title={G-CUT3R: Guided 3D Reconstruction with Camera and Depth Prior Integration},
  author={Khafizov, Ramil and Komarichev, Artem and Rakhimov, Ruslan and Wonka, Peter and Burnaev, Evgeny},
  journal={arXiv preprint arXiv:2508.11379},
  year={2025}
}

@article{mahdi2025evict3r,
  title={Evict3R: Training-Free Token Eviction for Memory-Bounded Streaming Visual Geometry Transformers},
  author={Mahdi, Soroush and Ayar, Fardin and Javanmardi, Ehsan and Tsukada, Manabu and Javanmardi, Mahdi},
  journal={arXiv preprint arXiv:2509.17650},
  year={2025}
}

@inproceedings{tang2025mvdust3r,
  title={Mv-dust3r+: Single-stage scene reconstruction from sparse views in 2 seconds},
  author={Tang, Zhenggang and Fan, Yuchen and Wang, Dilin and Xu, Hongyu and Ranjan, Rakesh and Schwing, Alexander and Yan, Zhicheng},
  booktitle=CVPR,
  pages={5283--5293},
  year={2025}
}

@inproceedings{elflein2025light3rsfm,
  title={Light3R-SfM: Towards Feed-forward Structure-from-Motion},
  author={Elflein, Sven and Zhou, Qunjie and Leal-Taix{\'e}, Laura},
  booktitle=CVPR,
  pages={16774--16784},
  year={2025}
}

@inproceedings{yang2025fast3r,
  title={Fast3r: Towards 3d reconstruction of 1000+ images in one forward pass},
  author={Yang, Jianing and Sax, Alexander and Liang, Kevin J and Henaff, Mikael and Tang, Hao and Cao, Ang and Chai, Joyce and Meier, Franziska and Feiszli, Matt},
  booktitle=CVPR,
  pages={21924--21935},
  year={2025}
}

@article{mahdavian2026uniscale,
  title={UniScale: Unified scale-aware 3D reconstruction for multi-view understanding via prior injection for robotic perception},
  author={Mahdavian, Mohammad and Tan, Gordon and Xu, Binbin and Ren, Yuan and Bai, Dongfeng and Liu, Bingbing},
  journal={arXiv preprint arXiv:2602.23224},
  year={2026}
}

@article{li2026cameras,
  title={Cameras as relative positional encoding},
  author={Li, Ruilong and Yi, Brent and Liu, Junchen and Gao, Hang and Ma, Yi and Kanazawa, Angjoo},
  journal={Advances in Neural Information Processing Systems},
  volume={38},
  pages={15984--16009},
  year={2026}
}

@inproceedings{miyato2024gta,
  title={Gta: A geometry-aware attention mechanism for multi-view transformers},
  author={Miyato, Takeru and Jaeger, Bernhard and Welling, Max and Geiger, Andreas},
  booktitle={International Conference on Learning Representations},
  volume={2024},
  pages={8172--8208},
  year={2024}
}

@article{cerezo2026efficient,
  title={An efficient closed-form solution to full visual-inertial state initialization},
  author={Cerezo, Samuel and Lee, Seong Hun and Civera, Javier},
  journal={IEEE Robotics and Automation Letters},
  year={2026},
  publisher={IEEE}
}

@article{qin2018vins,
  title={Vins-mono: A robust and versatile monocular visual-inertial state estimator},
  author={Qin, Tong and Li, Peiliang and Shen, Shaojie},
  journal={IEEE transactions on robotics},
  volume={34},
  number={4},
  pages={1004--1020},
  year={2018},
  publisher={IEEE}
}

@article{boche2025okvis2,
  title={OKVIS2-X: Open keyframe-based visual-inertial SLAM configurable with dense depth or LiDAR, and GNSS},
  author={Boche, Simon and Jung, Jaehyung and Laina, Sebasti{\'a}n Barbas and Leutenegger, Stefan},
  journal={IEEE Transactions on Robotics},
  year={2025},
  publisher={IEEE}
}

@inproceedings{veicht2024geocalib,
  title={Geocalib: Learning single-image calibration with geometric optimization},
  author={Veicht, Alexander and Sarlin, Paul-Edouard and Lindenberger, Philipp and Pollefeys, Marc},
  booktitle={European Conference on Computer Vision},
  pages={1--20},
  year={2024},
  organization={Springer}
}
